\documentclass[sigconf]{acmart}
\AtBeginDocument{%
  }

\renewcommand\footnotetextcopyrightpermission[1]{}
\newcommand{\mypara}[1]{\vspace{0.1cm}\noindent\textbf{#1}}

\renewcommand{\vec}[1]{\boldsymbol{#1}}

\usepackage{multirow}
\usepackage{lipsum}
\usepackage{graphicx}
\usepackage{marvosym}
\usepackage{balance}
\usepackage{fancyhdr}

\begin{document}

\title{Boosting Few-shot 3D Point Cloud Segmentation via Query-Guided Enhancement}


\author{
    Zhenhua Ning$^{1}$\footnotemark[2]
    \hspace{0.4cm} Zhuotao Tian$^{2}$\footnotemark[2]
    \hspace{0.4cm} Guangming Lu$^{1,3}$
    \hspace{0.4cm} Wenjie Pei$^{1}$\textsuperscript{\Letter}
}

\affiliation{$^{1}$
Harbin Institute of Technology~\city{Shenzhen}~\country{China}
\hspace{0.4cm} $^{2}$SmartMore~\country{China}\\ 
$^{3}$Guangdong Provincial Key Laboratory of Novel Security Intelligence Technologies
\vspace{0.5cm}}

\renewcommand{\shortauthors}{Zhenhua Ning et al.}

\begin{abstract}
Although extensive research has been conducted on 3D point cloud segmentation, effectively adapting generic models to novel categories remains a formidable challenge. This paper proposes a novel approach to improve point cloud few-shot segmentation (PC-FSS) models. Unlike existing PC-FSS methods that directly utilize categorical information from support prototypes to recognize novel classes in query samples, our method identifies two critical aspects that substantially enhance model performance by reducing contextual gaps between support prototypes and query features. Specifically, we (1) adapt support background prototypes to match query context while removing extraneous cues that may obscure foreground and background in query samples, and (2) holistically rectify support prototypes under the guidance of query features to emulate the latter having no semantic gap to the query targets. Our proposed designs are agnostic to the feature extractor, rendering them readily applicable to any prototype-based methods. The experimental results on S3DIS and ScanNet demonstrate notable practical benefits, as our approach achieves significant improvements while still maintaining high efficiency. The code for our approach is available at \href{https://github.com/AaronNZH/Boosting-Few-shot-3D-Point-Cloud-Segmentation-via-Query-Guided-Enhancement}{github.com/AaronNZH/Boosting-Few-shot-3D-Point-Cloud-Segmentation-via-Query-Guided-Enhancement}.
\end{abstract}

\begin{CCSXML}
<ccs2012>
   <concept>
       <concept_id>10010147</concept_id>
       <concept_desc>Computing methodologies</concept_desc>
       <concept_significance>500</concept_significance>
       </concept>
   <concept>
       <concept_id>10010147.10010178</concept_id>
       <concept_desc>Computing methodologies~Artificial intelligence</concept_desc>
       <concept_significance>500</concept_significance>
       </concept>
   <concept>
       <concept_id>10010147.10010178.10010224</concept_id>
       <concept_desc>Computing methodologies~Computer vision</concept_desc>
       <concept_significance>500</concept_significance>
       </concept>
   <concept>
       <concept_id>10010147.10010178.10010224.10010225</concept_id>
       <concept_desc>Computing methodologies~Computer vision tasks</concept_desc>
       <concept_significance>500</concept_significance>
       </concept>
   <concept>
       <concept_id>10010147.10010178.10010224.10010225.10010227</concept_id>
       <concept_desc>Computing methodologies~Scene understanding</concept_desc>
       <concept_significance>500</concept_significance>
       </concept>
 </ccs2012>
\end{CCSXML}

\ccsdesc[500]{Computing methodologies}
\ccsdesc[500]{Computing methodologies~Artificial intelligence}
\ccsdesc[500]{Computing methodologies~Computer vision}
\ccsdesc[500]{Computing methodologies~Computer vision tasks}
\ccsdesc[500]{Computing methodologies~Scene understanding}

\keywords{representation learning, 3D scene understanding, few-shot segmentation}

\maketitle

\renewcommand{\thefootnote}{\fnsymbol{footnote}}
\footnotetext[2]{These authors contributed equally to this work. \\
\textsuperscript{\Letter} Corresponding author: \textit{W. Pei} (wenjiecoder@outlook.com)}

\begin{figure}[t]
\centering
\includegraphics[width=\linewidth]{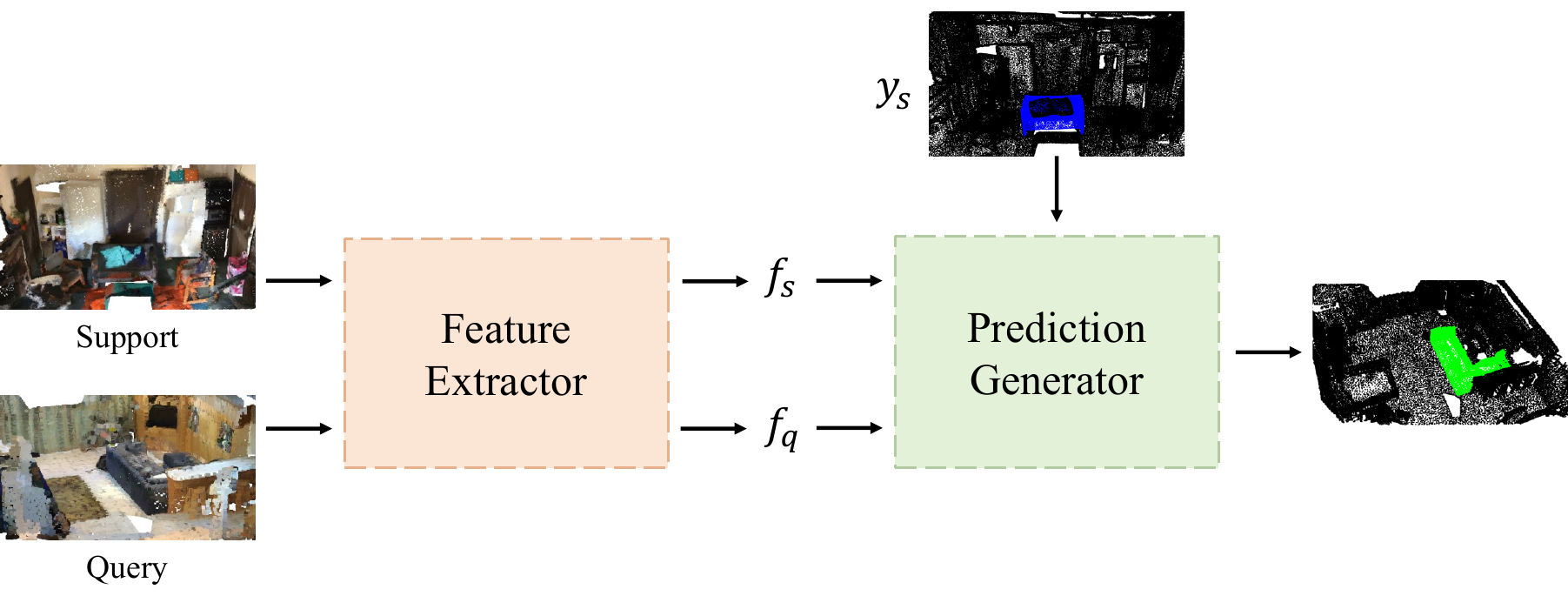}
\caption{PC-FSS aims to identify previously unseen target areas in a query point cloud using categorical information provided by a support sample and the corresponding mask $\mathcal{Y}_s$ (1-way 1-shot in this figure). The extracted query and support features are denoted as $\vec{f}_q$ and $\vec{f}_s$, respectively.}
\label{fig:FSS}
\vspace{-0.3cm}
\end{figure}

\section{Introduction}
\label{ref:intro}

Semantic segmentation constitutes a fundamental albeit challenging task for scene parsing, demanding the generation of precise pixel-wise predictions for each object within the scene.
Recently, it has attracted significant attention~\cite{choy20194d,efficient3d,jiang2021guided,zhao2019pointweb,ao2020SpinNet,tian2023cac,deng2020voxel,rr} due to its potential applications in a variety of fields, including autonomous driving~\cite{fastseg_autodrive}, robot navigation~\cite{robot,3dsemiseg}, and many more~\cite{mediseg,stratified,shapeawareembedding,semiseg,decouplenet}. 

Nonetheless, once the generic frameworks are trained, their ability to handle previously unseen classes in novel tasks may be impeded by the lack of sufficient fully-labeled data. Even if the requisite data for new classes is available, careful fine-tuning incurs additional costs in terms of time and resources. In order to make the model learn to rapidly adapt to previously unseen classes, point-cloud few-shot segmentation (PC-FSS) has been proposed~\cite{zhao2021few}. As depicted in Figure \ref{fig:FSS}, the prediction generator utilizes the categorical information mined from the support samples to generate predictions for query samples. In essence, this approach mimics the process by which a model identifies targets from previously unseen classes in testing point clouds (query), with the help of a limited number of annotations provided by the reference (support).

\begin{figure*}[t]
  \includegraphics[width=\linewidth]{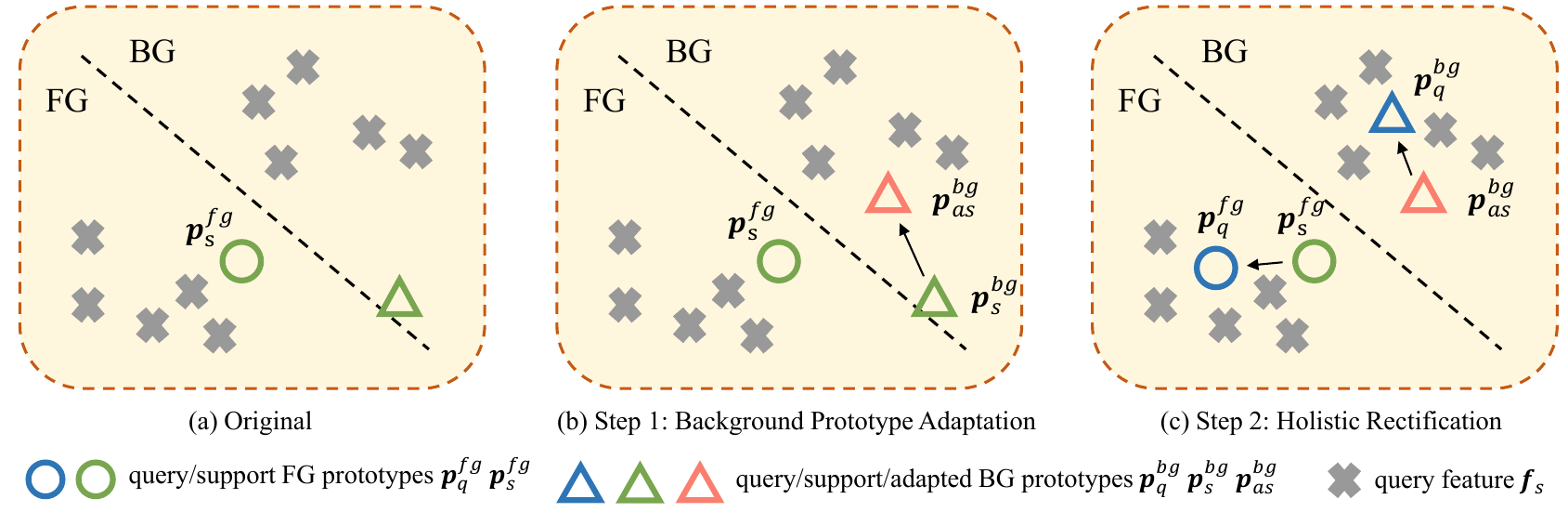}
  \caption{The workflow of our method. In Step 1, support background prototypes $\vec{p}_s^{bg}$ are adapted to query background context as $\vec{p_{as}^{bg}}$. In Step 2, query prototypes work as teachers to rectify support prototypes holistically.}
  \label{fig:teaser}
  \vspace{0.1cm}
\end{figure*}

However, existing methods~\cite{prototype_cls,zhao2021few} fail to account for the fact that the query and support point clouds always contain different contents, which could result in categorical representations being biased toward the contextual cues of the support samples. This limits their ability to function as reliable and comprehensive class descriptors for object identification in query point clouds with diverse scene variations for the target classes. Therefore, in order to tackle the semantic gap between query features and support prototypes, we have identified two key steps that need to be addressed as shown in Figure~\ref{fig:teaser}. 

\textbf{Step 1: Adapting support background prototypes to fit the query context. }
As an example, consider a scenario with only one novel class, where the objective is to identify the foreground (FG) region belonging to the target class provided by the support set and label the remaining areas on the query point clouds as the background (BG). However, to distinguish between the FG and BG regions on the query point clouds, current approaches often use mask-pooling to extract the support FG and BG prototypes. While the query and support FG regions share the same categorical information, the BG regions may vary significantly due to scene changes, and the support BG may contain objects with similar geometric features to the query FG targets. Hence, refining the BG prototypical representations is imperative to encourage the support BG prototypes to be better aligned with the query BG, and be devoid of query FG target-related cues that could potentially result in erroneous predictions. 

\textbf{Step 2: Optimizing support prototypes to holistically suit the query context. }
The preceding step refines the support BG prototypes while leaving the FG prototypes intact. However, as detailed in Section~\ref{sec:method}, we have empirically observed a discrepancy between the support FG prototypes and the query FG prototypes due to contextual differences. To address this, during training, we utilize the query FG/BG prototypes as a holistic guide for the corresponding support prototypes. This approach enables the support prototypes to learn to replicate the performance of the superior categorical representations directly obtained from the query itself, thus mitigating the adverse effects stemming from the semantic gaps between the query and support point clouds.

To overcome these challenges, we present two innovative query-guided strategies: Background Prototype Adaptation (BPA) and Holistic Rectification (HR). Our approaches are independent of feature extraction and can be seamlessly integrated into any prototype-based PC-FSS framework. In essence, our contributions can be summarized in three points:
\begin{itemize}
    \item Our study reveals the impact of semantic gaps between query and support samples, which has been overlooked by previous methods and may hinder the performance of PC-FSS.
    \item To address this challenge, we propose two novel and efficient strategies, namely Background Prototype Adaptation (BPA) and Holistic Rectification (HR).
    \item Our approaches are agnostic to the feature extractor and can be easily incorporated into different PC-FSS frameworks, resulting in significant performance gains without compromising model efficiency.
\end{itemize}

\section{Preliminaries}
\label{sec:preliminary}
This section provides the necessary preliminaries before introducing our method. The introduction to related works in the broader field is presented in Section~\ref{sec:related_work}.

\mypara{Task description. }
The objective of point cloud few-shot segmentation (PC-FSS) is to enable rapid adaptation of models to newly encountered categories. Evaluation of PC-FSS is conducted episodically, with each episode comprising a support set $S$ and a query set $Q$. The support set $S$, consisting of a small number of annotated samples, provides information on the novel categories, which the models must then utilize to identify the corresponding regions within the query samples.

To be more specific, as per the investigation conducted in \cite{zhao2021few}, the $N$-way $K$-shot PC-FSS problem is considered, whereby the support set comprises $N$ newly encountered categories, each of which is represented by $K$ annotated samples.
The categories are partitioned into two mutually exclusive sets, namely $C_{tr}$ and $C_{te}$, with the former being utilized for model training and the latter being utilized for evaluating the model's ability to generalize to previously unseen categories.

In each $N$-way $K$-shot episode, a query set $Q = \{\vec{x}_q^{t},~\vec{y}_q^{t}\}_{t=1}^T$ is provided, comprising $T$ pairs of query point clouds $\vec{x}_q^{t}$ and their corresponding labels $\vec{y}_q^{t}$, along with $S = \{\{\vec{x}_{s}^{n,k},~\vec{y}_{s}^{n,k}\}_{k=1}^{K}\}_{n=1}^N$ comprising $K$ pairs of support point clouds $\vec{x}_{s}^{n,k}$ and their corresponding labels $\vec{y}_{s}^{n,k}$ for each of $N$ categories. We note that  both the query labels $\vec{y}_{q}^{t}$ and the support labels $\vec{y}_{s}^{n,k}$ are utilized during the training phase, while only the support labels $\vec{y}_{s}^{n,k}$ are available during testing for imparting novel categorical information. Models are required to identify the target classes in $Q$ based on the categorical information provided by $S$.

In what follows, we provide a brief introduction to two closely related methods~\cite{prototype_cls,zhao2021few}, which serve as our baseline models as well. Both approaches follow the meta-structure depicted in Figure~\ref{fig:FSS}, but employ distinct strategies for prediction generation.

\mypara{ProtoNet. }
It~\cite{prototype_cls} was originally proposed to address the few-shot image classification problem and was subsequently adapted for 3D point cloud segmentation by~\cite{zhao2021few}. ProtoNet employs DGCNN with a linear projector as the feature extractor to generate the support feature $\vec{f}_s$ and query feature $\vec{f}_q$. Global average pooling is performed upon the support feature map to obtain a prototype for each category. For yielding predictions, ProtoNet computes the cosine similarity between the query feature and the prototypes to produce the final results. The similar idea has been adopted in few-shot image segmentation~\cite{wang2019panet}.

\mypara{AttMPTI. }
Different from ProtoNet, AttMPTI~\cite{zhao2021few} utilizes a DGCNN with a self-attention layer as the feature extractor. Moreover, AttMPTI adopts a distinct approach for generating multi-prototypes, which involves sampling seed points from support points and assigning each point to the closest seed based on the distance in the feature space. Subsequently, using the multi-prototypes and query feature as nodes, AttMPTI constructs a $k$-NN graph~\cite{llm_kg,liang2023knowledge} and applies label propagation \cite{iscen2019label} to obtain the final predictions. AttMPTI is the latest state-of-the-art method.

\begin{figure}[!t]
  \centering
  \includegraphics[width=0.7\linewidth, height=0.5\linewidth]{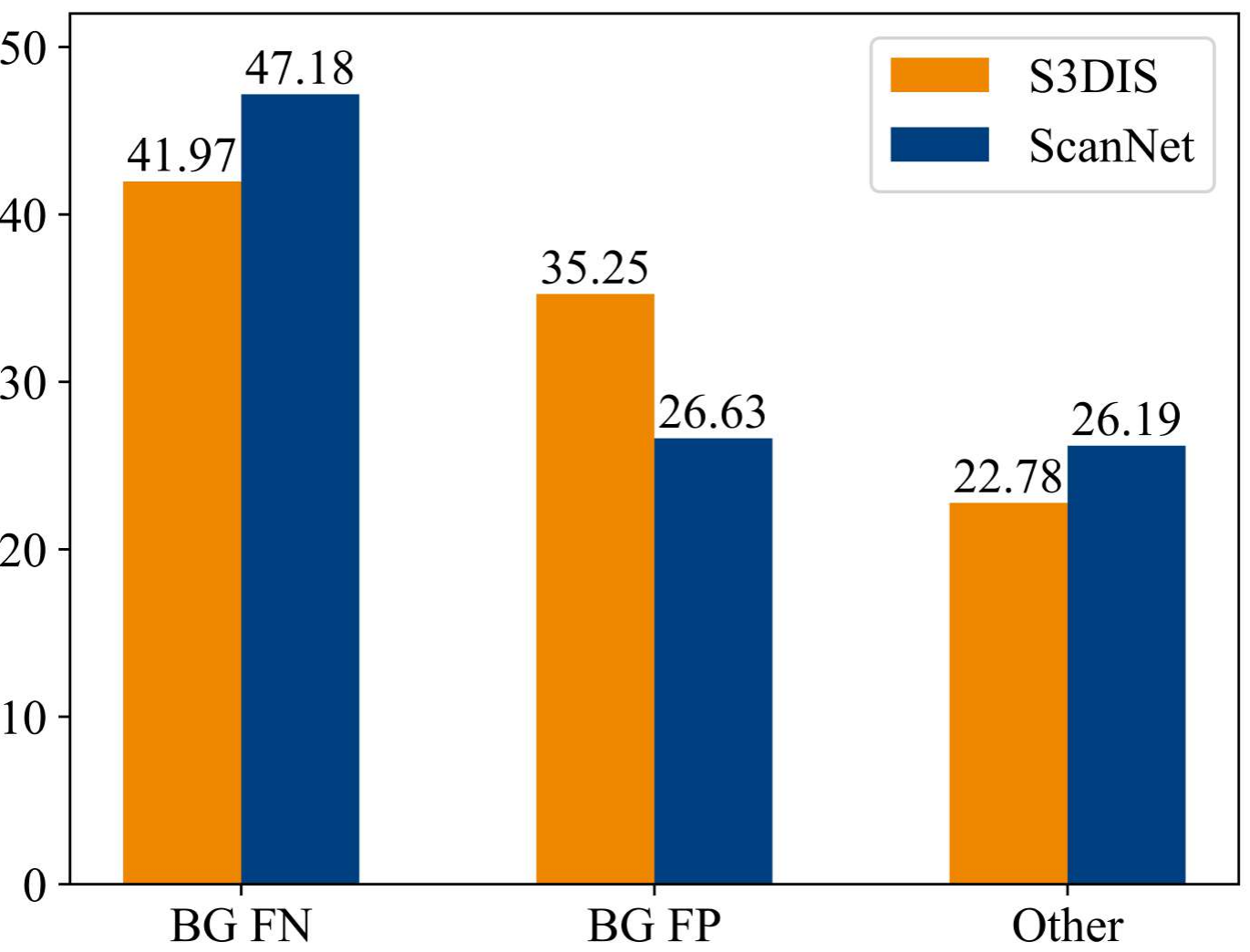}
  \caption{The statistics of AttMPTI's prediction errors on S3DIS and ScanNet, under the 2-way 1-shot test setting.}
  \Description{}
  \label{fig:statistics_of_pred_errors}
\end{figure}

\section{Our Approach}
\label{sec:method}
In the following, 
Sections~\ref{sec:bpa} and \ref{sec:hr} elaborate on the technical details of the proposed Background Prototype Adaptation (BPA) and Holistic Rectification (HR), respectively, while also highlighting the key observations and motivations that guided their development. Then, in Section~\ref{sec:overview}, we provide an overview of how these techniques can be integrated into our baseline models.

\subsection{Background Prototype Adaptation}
\label{sec:bpa}

\mypara{Observation. } 
Despite the impressive performance demonstrated by the prior work \cite{zhao2021few}, it neglects the potential impact of variations in the background (BG) between the support and query sets, which can lead to a substantial number of false negatives in the prediction of background points, as exemplified in Figure \ref{fig:statistics_of_pred_errors}. We hypothesize that this may be attributed to the fact that the support background prototypes $\vec{p}_s^{bg}$ are insufficient to differentiate between the foreground and background points in the query set.

\begin{figure}[!t]
  \centering
  \includegraphics[width=\linewidth]{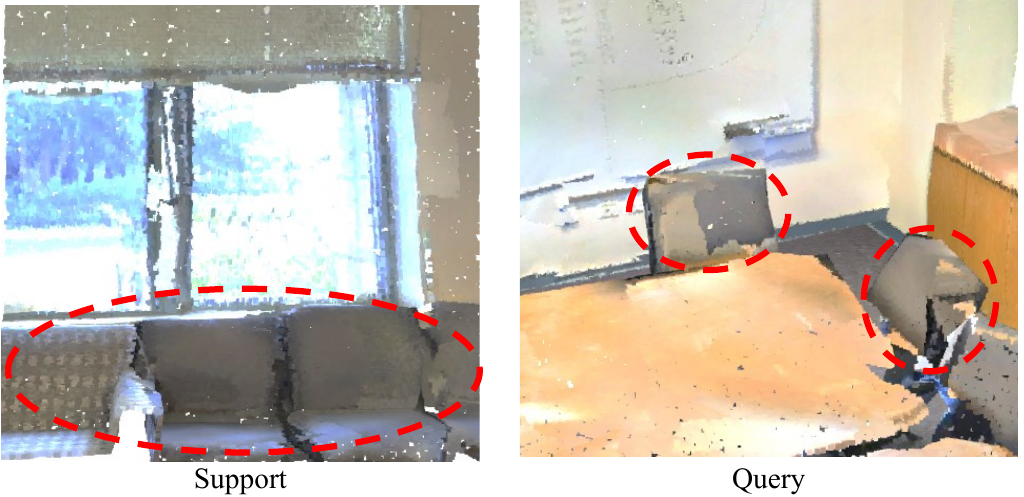}
  \caption{Visualizations of two point clouds of S3DIS, with the image on the left representing the support and the other representing the query. The chairs marked by red circles in both images are foreground targets, while the remaining regions are classified as background.}
  \Description{}
  \label{fig:background_compare}
\end{figure}

To illustrate our point, take Figure \ref{fig:background_compare} as an example, where the chair is considered as the foreground and the remaining objects constitute the background. It is evident that the backgrounds of the support and query sets are dissimilar, with the former comprising mostly windows and the latter consisting of tables and boards. Consequently, the support background prototypes $\vec{p}_s^{bg}$ may not be able to effectively represent the background of the query set.
Moreover, given that tables exhibit a greater degree of partial similarity with chairs as opposed to ceilings and floors, the table in the query set is more prone to being misidentified as foreground. Accordingly, it is crucial to \textit{boost the discriminative capability of the support background prototypes $\vec{p}_s^{bg}$ by adapting $\vec{p}_s^{bg}$ to the query context but eliminating information that is relevant to the foreground of the query set.}
% \zttian{add visualized predictions to support your claim. can do it later}

\mypara{Our method. } 
% A natural approach to tackle this challenge would be to leverage the correlation between the foreground in the support and query sets, by enabling $\vec{p}_s^{fg}$ to interact with $\vec{f}_s$ and $\vec{p}_s^{bg}$. Building on this idea, we propose an innovative method that can incorporate information from $\vec{f}_s$ while simultaneously filtering out the foreground-related information in $\vec{p}_s^{fg}$, thereby enhancing the discriminative power of $\vec{p}_s^{bg}$. Remarkably, our initial attempts based on this approach yielded significantly improved results.
The meta-view of the approach to tackle the aforementioned challenge is: 1) establishing a module $\mathcal{C}$ to find the correlation $\mathcal{r}$ between the support background prototypes $\vec{p}_s^{bg}$ and query features $\vec{f}_q$, then 2) yielding adapted support background prototypes $\vec{p}_{as}^{bg}$ conditioned on $\mathcal{r}$ via a rectification module $\mathcal{R}$:
\begin{equation}
    \label{eq:proto_adapt}
    \mathcal{r} = \mathcal{C}(\vec{p}_s^{bg}, \vec{f}_q) \quad
    \vec{p}_{as}^{bg} = \mathcal{R}(\vec{p}_s^{bg}, \mathcal{r}).
\end{equation}
A straightforward choice for $\mathcal{C}$ is a cross-attention layer that directly computes the correlation $\mathcal{r}$ between $\vec{p}_s^{bg}$ and $\vec{f}_q$. However,  this approach results in a noticeable reduction in performance compared to the baseline as shown in Section~\ref{sec:ablation_study}. This can be attributed to the fact that cross-attention only produces the content that is correlated, which could inadvertently incorporate query foreground cues into $\vec{p}_{as}^{bg}$, thereby exacerbating the challenge of distinguishing between the foreground and background points.

Alternatively, we may screen out the query foreground-related hints from $\mathcal{r}$ with the aid of support foreground prototypes $\vec{p}_s^{fg}$ that can be accurately produced with the support masks. Thus, $\mathcal{r}$ can be yielded as:
\begin{equation}
    \mathcal{r} = \mathcal{C}(\vec{p}_s^{bg}, \vec{f}_q, \vec{p}_s^{fg}).
\end{equation}
To address this issue, one may opt to use a learnable module for $\mathcal{C}$, which can learn to inherently estimate the correlations between the inputs and output $\mathcal{r}$. However, as presented in the ablation study, we observed that simultaneously modeling both the \textit{exclusive relationship}  between $\vec{p}_s^{bg}$ and $\vec{p}_s^{fg}$, as well as the \textit{inclusive relationship}  between $\vec{p}_s^{bg}$ and $\vec{f}_q$ within a shared module, is challenging and can lead to further performance deterioration. Moreover, since the point numbers of $\vec{f}_q$ and $\vec{p}_s$ may differ, using a learnable module may not be compatible with the approach based on multiple prototypes. Therefore, we opted to explicitly and independently model these relationships.

To be specific, we adopt module $\mathcal{C}_1$ to estimate the inclusive relationship $\mathcal{r}_q$ from $\vec{p}_s^{bg}$ and $\vec{f}_q$, and adopt another module $\mathcal{C}_2$ to yield the exclusive relationship  $\mathcal{r}_s$ from $\vec{p}_s^{bg}$ and $\vec{p}_s^{fg}$, which can be formulated as:
\begin{equation}
    \mathcal{r}_q = \mathcal{C}_1(\vec{p}_s^{bg}, \vec{f}_q) \quad
    \mathcal{r}_s = \mathcal{C}_2(\vec{p}_s^{bg}, \vec{p}_s^{fg}).
\label{eq:query_adaptation}
\end{equation}
Then, the query background context $\mathcal{r}$ used for adapting $\vec{p}_s^{bg}$ can be approximated by $\mathcal{r}_q$ and $\mathcal{r}_s$:
\begin{equation}
    \mathcal{r} = \mathcal{G}( \mathcal{r}_q,  \mathcal{r}_s) = \mathcal{r}_q \cdot ( 1 - \sigma(\mathcal{r}_s)).
\label{eq:FG_information_filter}
\end{equation}
Here, $\sigma$ represents an MLP module followed by a \textit{sigmoid} function. By incorporating this into $\mathcal{G}$, the foreground target-related features are removed via $(1 - \sigma(\mathcal{r}_s))$, while the query background context is still retained in $\mathcal{r}_q$. Finally, $\mathcal{r}$ is used to yield the adapted support background prototypes $\vec{p}_{as}^{bg}$ with rectification module $\mathcal{R}$ via $\vec{p}_{as}^{bg} = \mathcal{R}(\vec{p}_s^{bg}, \mathcal{r})$ shown in Eq.~\eqref{eq:proto_adapt}. 

The approach seems to be applicable to adapt the support foreground prototypes $\vec{p}_s^{fg}$, i.e., $\mathcal{r}_q = \mathcal{C}_1(\vec{p}_s^{fg}, \vec{f}_q)$, in addition to the background. However, we observed that this does not lead to further improvements, as demonstrated in our ablation study in Section~\ref{sec:ablation_study}. 
The challenge of estimating $\mathcal{r}$, in this case, can be explained by the fact that $\vec{p}_s^{bg}$ in $\mathcal{r}_s = \mathcal{C}_2(\vec{p}_s^{fg}, \vec{p}_s^{bg})$ may not be able to represent the background of the query, thus it fails to  screen out the background for query targets.

Both $\mathcal{C}_1$, $\mathcal{C}_2$ and $\mathcal{R}$ are implemented as single-layer cross-attention modules, while $\sigma$ is realized as a 2-layer MLP followed by a $sigmoid$ function. It is important to mention that there might exist more complex designs for these modules, but in this study, we do not focus on investigating advanced architectures for these components.

\begin{figure*}[t]
\centering
\includegraphics[width=\textwidth]{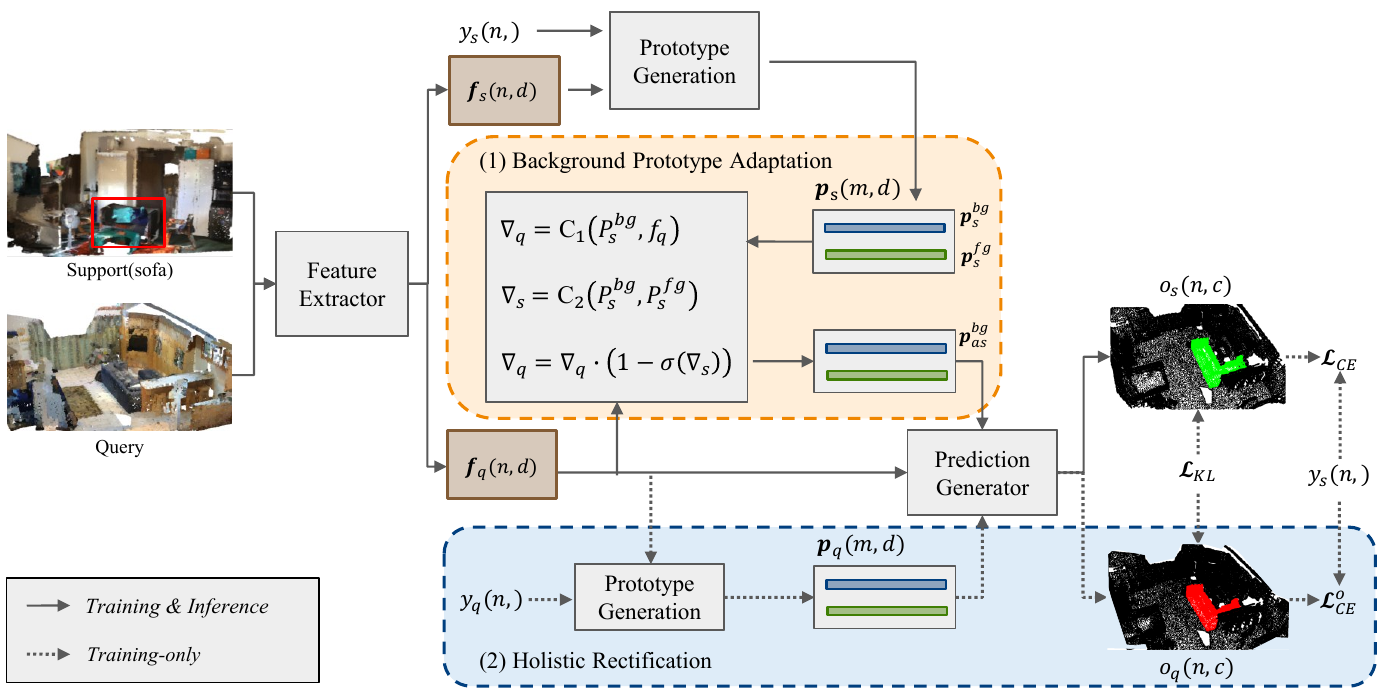}
\caption{The framework of our method. There are four key parameters: $n$, $m$, $d$, and $c$, which respectively represent the number of points, prototypes, feature dimension channels, and categories. Additionally, the Prediction Generator can be implemented in one of two ways: either through the use of the $k$-NN Graph in AttMPTI, or by employing cosine similarity calculations in ProtoNet.}
\Description{}
\label{fig:framwork_overview}
\vspace{0.1cm}
\end{figure*}

\subsection{Holistic Rectification}
\label{sec:hr}

\begin{table}[!t]
\tabcolsep=1.0cm
\begin{tabular}{lc}
\toprule
Methods & mIoU(\%) \\
\midrule
Support Prototypes & 66.40        \\
Optimal Query Prototypes   & 93.89        \\
\bottomrule
\end{tabular}
\caption{Comparison of using support prototypes and the optimal query prototypes on $S^0$ of S3DIS under the 1-way 1-shot  setting. The optimal query prototypes are obtained via query ground truth during training.}
\label{tab:case_study}
\vspace{-0.3cm}
\end{table}

\mypara{Another challenge. }The aforementioned BPA helps to mitigate contextual inconsistencies between the query and support background regions. However, since BPA is not applicable to the support foreground prototypes, the presence of distinct objects in query and support point clouds may still give rise to undesired semantic gaps between support foreground prototypes $\vec{P}_{s}^{fg}$ and the target regions of $\vec{f}_q$, potentially resulting in suboptimal performance when attempting to locate target regions in query samples.

To examine our hypothesis that the existence of diverse objects in both query and support point clouds may negatively impact foreground representations, we replaced the support prototypes with the ones derived from query labels and features, i.e., $\vec{p}_s \coloneqq \vec{p}_q$. As illustrated in Table~\ref{tab:case_study}, a significant discrepancy was observed between the results obtained using support prototypes and optimal query prototypes, lending support to our hypothesis. Addressing the performance gap is the subsequent crucial step.

\mypara{Inspiration from knowledge distillation. } As the query labels are only available during training, we are unable to replace the support prototypes with the optimal prototypes obtained using query labels for testing. 

Motivated by the concept of knowledge distillation~\cite{hinton2015distilling,decouplekld,apd}, which entails transferring latent knowledge from a superior representation (teacher) to a weaker one (student) via minimizing the KL divergence, we seek to extract valuable semantic information from the optimal prototypes $\vec{p}_q$ derived from the query feature itself and inject them into the support prototypes $\vec{p}_s$. This approach aims to facilitate the model in learning how to produce support prototypes that align more closely with the query context by emulating the behavior of the optimal query prototypes.

\mypara{Learn from the query guidance. }
To simplify matters, we take the 1-way 1-shot task as an example in the following, where there is only one target class and one prototype for each category by global average mask pooling. In this case, $\vec{y}_{q} \in \Omega^{[n \times 1]}$ is a binary mask for the input query point cloud with $n$ elements. Specifically, we first get the optimal query prototype $\vec{p}_q \in \Omega^{[2 \times d]}$, constituting the fore- and background prototypes ($\vec{p}_q = [\vec{p}_q^{bg}; \vec{p}_q^{fg}]$), via mask pooling the extracted query feature $\vec{f}_{q} \in \Omega^{[n \times d]}$ with the binary query mask $\vec{y}_{q} \in \Omega^{[n \times 1]}$:
\begin{equation}
    \vec{p}_q^{fg} = \frac{1}{n}\sum_{i=1}^{n} \vec{f}_q^i \cdot \vec{y}_q^i 
    \quad
    \vec{p}_q^{bg} = \frac{1}{n}\sum_{i=1}^{n} \vec{f}_q^i \cdot (1-\vec{y}_q^i).
\end{equation}
Then, we can get the predictions $\vec{o}_q \in \Omega^{[n \times 2]}$ and $\vec{o}_s \in \Omega^{[n \times 2]}$ from $\vec{p}_q$ and $\vec{p}_s$ respectively by measuring their cosine similarities with $\vec{f}_{q} \in \Omega^{[m \times d]}$:
\begin{equation}
    \vec{o}_q = \vec{t} \cdot \theta(\vec{f}_{q}, \vec{p}_q) 
    \qquad
    \vec{o}_s = \vec{t} \cdot \theta(\vec{f}_{q}, \vec{p}_s).
\end{equation}
Here, $\vec{t}$ represents a temperature used to scale the value range of cosine $\theta$ from $[-1,1]$ to $[-\vec{t}, \vec{t}]$,  such that the cosine outputs can be decently optimized by the cross-entropy loss. Following \cite{wang2019panet}, we empirically set $\vec{t}$ to 15. The sensitivity analysis regarding $\vec{t}$ is shown in the appendix.

Subsequently, we minimize the KL divergence to promote similarity between $\vec{o}_s$ and $\vec{o}_q$:
\begin{equation}
  \mathcal{L}_{KL} = \frac{1}{n} \sum_{i=1}^{n}KLD(\vec{o}_s^i \| \vec{o}_q^i).
  \label{KL loss}
\end{equation}
To prevent the model from collapsing into trivial cases, such as producing all-same features, the gradients of $\vec{o}_q$ obtained from $\mathcal{L}_{KL}$ are detached during back-propagation. 

While the direct prototype alignment through minimizing the L1/L2 losses may be a feasible approach, our experiments in Section~\ref{sec:ablation_study} demonstrate that minimizing KL divergence between predicted logits leads to superior results. This may be attributed to the fact that KL divergence can expose more hidden knowledge, as suggested in \cite{hinton2015distilling}.

It is worth noting that the proposed distillation method is architecture agnostic, allowing for straightforward adaptation to other few-shot prototype-based approaches. Besides, this process is exclusively implemented during training and does not impose any additional computational burden on inference. 

\subsection{Integration}
\label{sec:overview}
\mypara{Structural overview. }
The two aforementioned designs serve the purpose of 1) reducing the semantic gaps between the support background prototypes and query background features, and 2) holistically aligning the semantic contexts between query features and support prototypes. This section furnishes an elaborate description of how these designs are integrated into the baseline models.

The overview of the framework incorporated with our proposed methods is shown in Figure \ref{fig:framwork_overview}. As different models may adopt distinct designs for generating support prototypes and making final predictions, we present with abstractions in the figure, i.e., two modules named `Prototype Generation' and `Prediction Generation' respectively.

To elaborate on the process, the Feature Extractor is first used to extract features $\vec{f}_s$ and $\vec{f}_q$ from the support and query points, respectively. Next, the Prototype Generation module generates the raw support prototypes $\vec{p}_s$ by combining $\vec{f}_s$ and support labels $\vec{y}_s$. However, as highlighted in Section~\ref{sec:bpa}, the misalignment between support and query backgrounds can introduce undesirable information that degrades the performance. Therefore, we apply the proposed Background Prototype Adaptation (BPA) to $\vec{p}_s$ to obtain the query-aligned background prototype in $\vec{p}_{as}$, conditioned on the query features $\vec{f}_q$.

Besides, as demonstrated by the findings in Table~\ref{tab:case_study} in Section~\ref{sec:hr}, the presence of diverse co-occurring objects can also impact the foreground representations, resulting in a semantic mismatch between the support foreground prototype and the corresponding query regions.
To address this issue, we also employ the proposed Holistic Rectification strategy to further refine the support prototypes $\vec{p}_{as}$ obtained after the BPA part. This is achieved by encouraging $\vec{p}_{as}=[\vec{p}_{as}^{bg};\vec{p}_{as}^{fg}]$ to resemble $\vec{p}_q=[\vec{p}_q^{bg};\vec{p}_q^{fg}]$ that better aligns with the query's context, thus minimizing the semantic discordance between the support foreground prototypes and the query targets.

Both the two efficient designs do not have any specific structural requirements for the feature extractor, and the Holistic Rectification is only used during training. Therefore, they can be easily integrated into various prototype-based PC-FSS frameworks without sacrificing overall efficiency.

\mypara{Training objective. }
For the baseline model without our proposed designs, the training loss is computed as the cross-entropy loss $\mathcal{L}_{ce}$ applied to the query predictions obtained with the support prototypes.

However, our proposed Holistic Rectification (HR)  encourages the support prototypes to emulate the optimal query prototypes, which necessitates the addition of the Kullback-Leibler (KL) loss $\mathcal{L}_{KL}$. Furthermore, since the optimal query prototypes serve as the teacher during the distillation process, it is important to ensure their validity. Therefore, we additionally incorporate the cross-entropy loss $\mathcal{L}_{ce}^{o}$ to supervise the predictions generated by the optimal query prototypes, and the overall training objective $\mathcal{L}$ is defined as the sum of these three losses:
\begin{equation}
    \mathcal{L} = \mathcal{L}_{ce} + \mathcal{L}_{ce}^o + \lambda_{KL}  \mathcal{L}_{KL}.
\end{equation}
The weight for the KL loss $\mathcal{L}_{KL}$ is denoted by $\lambda_{KL}$. We empirically set this hyperparameter to 1, and our ablation study in Section~\ref{sec:ablation_study} demonstrates that this choice leads to satisfactory performance.

\begin{figure}[!t]
\centering
\includegraphics[width=\linewidth]{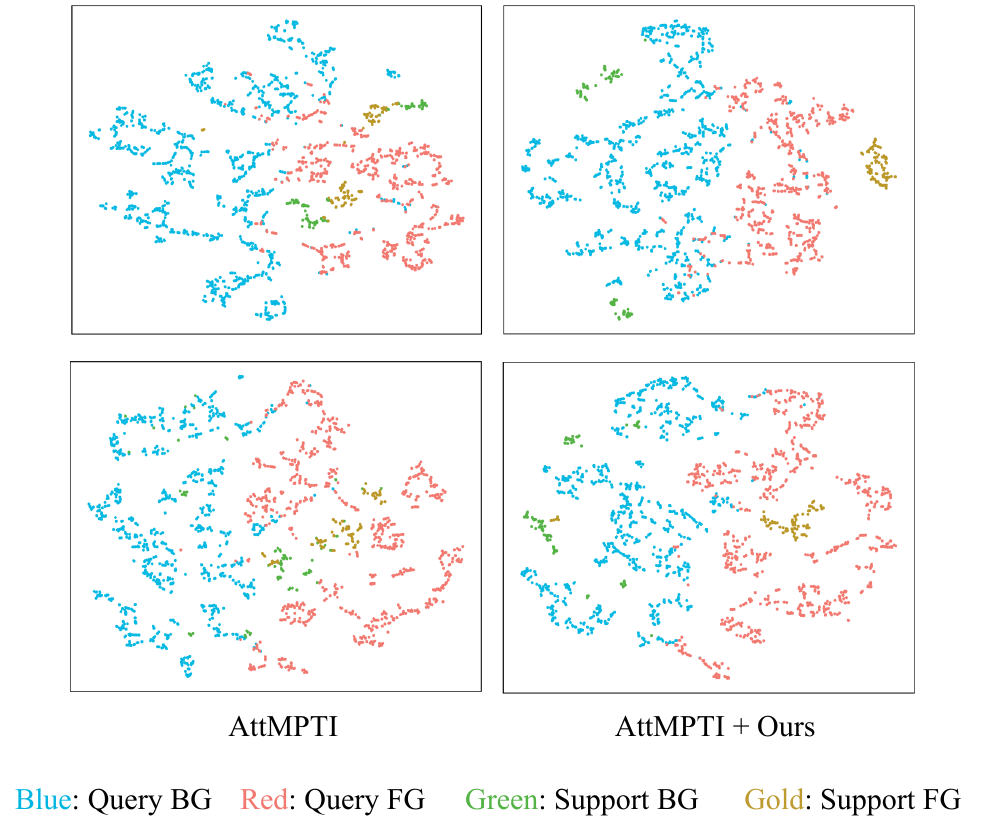}
\caption{The t-SNE results are presented with different colors representing the query and support prototypes. Specifically, the query background and foreground features are denoted by blue and red, respectively, while the support background and foreground prototypes are depicted using green and gold.}
\label{fig:tsne1}
\vspace{0.1cm}
\end{figure}

\begin{table}[t]
\tabcolsep=0.15cm
\begin{tabular}{lcccc}
\toprule
                            & \multicolumn{1}{c}{1-way 1-shot} & \multicolumn{1}{c}{1-way 5-shot} & \multicolumn{1}{c}{2-way 1-shot} & \multicolumn{1}{c}{2-way 5-shot} \\
\midrule
\multicolumn{1}{c}{$n_{q}$} & 100                         & 100                         & 100                       & 100  \\
\multicolumn{1}{c}{$n_{s}$} & 100                         & 100                         & 100                       & 150  \\         
\bottomrule
\end{tabular}
\caption{The numbers of prototypes in different settings. $n_{q}$ and $n_{s}$ denote the numbers of query and support prototypes respectively.}
\label{tab:n_prototypes}
\vspace{-0.3cm}
\end{table}

\section{EXPERIMENTS}

\subsection{Datasets and Setup}
In accordance with AttMPTI~\cite{zhao2021few}, our method is assessed using the S3DIS~\cite{2017arXiv170201105A} and ScanNet~\cite{dai2017scannet} datasets. The classes within each dataset are partitioned into two non-overlapping subsets, denoted as $S^{0}$ and $S^{1}$ for being utilized as mutual validation sets, with one subset designated as the test class $C_{te}$ and the other as the train class $C_{tr}$. Please refer to the appendix for further details regarding the partitioning process. Our approach employs the same data pre-processing strategy used in the prior work~\cite{zhao2021few}, which divides all rooms into blocks with dimensions of 1m x 1m. Each block is then randomly sampled to contain 2048 points.

\subsection{Implementation Details}
\mypara{Training.} Models are implemented using PyTorch and trained on an NVIDIA GeForce RTX 3090 GPU. The DGCNN-based feature extractor module, followed by three MLP layers, is pre-trained on the training set for 100 epochs with a batch size of 32. To optimize the pre-trained model, we use Adam with a learning rate of 0.001. During episodic training, we utilize the pre-trained weights to initialize the feature extractor and set its initial learning rate to 0.0001, and 0.001 for all remaining modules in the model. The learning rates are halved every 5000 iterations. Additionally, the batch size is set to 1, and Adam is adopted as the optimizer. Models are run for 40,000 iterations, following \cite{zhao2021few}. In each iteration, we randomly sample an episode whose category belongs to $C_{tr}$ as the data and utilize Gaussian jittering and random rotation around the z-axis as the data augmentation.

\mypara{Hyperparameters.} There are two additional hyperparameters $i.e.$ $n_q$ and $n_s$ when applying our method on AttMPTI~\cite{zhao2021few} that adopts multiple prototypes, as shown in Table \ref{tab:n_prototypes}. Following \cite{zhao2021few}, we set different values for the numbers of $\vec{p}_q$ and $\vec{p}_s$, namely $n_q$ and $n_s$, under different n-way k-shot settings. Notably, $n_q$ is shot agnostic, since the numbers of query point clouds are the same in different k-shot cases. All remaining hyperparameters are the same as \cite{zhao2021few}.

\subsection{Results and Analyses}

\begin{table*}[!t]
\centering
\begin{tabular}{c|cccccc|cccccc}
\toprule
  \multicolumn{1}{c|}{\multirow{3}{*}{Method}} &
  \multicolumn{6}{c|}{\textbf{1-way}} &
  \multicolumn{6}{c}{\textbf{2-way}} \\
  \cline{2-13} 
 &
  \multicolumn{3}{c|}{\textbf{1-shot}} &
  \multicolumn{3}{c|}{\textbf{5-shot}} &
  \multicolumn{3}{c|}{\textbf{1-shot}} &
  \multicolumn{3}{c}{\textbf{5-shot}} \\
  \cline{2-13} 
 &
  \multicolumn{1}{c|}{S0} &
  \multicolumn{1}{c|}{S1} &
  \multicolumn{1}{c|}{Mean} &
  \multicolumn{1}{c|}{S0} &
  \multicolumn{1}{c|}{S1} &
  Mean &
  \multicolumn{1}{c|}{S0} &
  \multicolumn{1}{c|}{S1} &
  \multicolumn{1}{c|}{Mean} &
  \multicolumn{1}{c|}{S0} &
  \multicolumn{1}{c|}{S1} &
  Mean \\
\hline
\multicolumn{13}{c}{S3DIS} \\ \hline
ProtoNet~\cite{prototype_cls} &
  \multicolumn{1}{c|}{66.18} &
  \multicolumn{1}{c|}{67.05} &
  \multicolumn{1}{c|}{66.62} &
  \multicolumn{1}{c|}{70.63} &
  \multicolumn{1}{c|}{72.46} &
  71.55 &
  \multicolumn{1}{c|}{48.39} &
  \multicolumn{1}{c|}{49.98} &
  \multicolumn{1}{c|}{49.19} &
  \multicolumn{1}{c|}{57.34} &
  \multicolumn{1}{c|}{63.22} &
  60.28 \\ \hline
AttMPTI \cite{zhao2021few} &
  \multicolumn{1}{c|}{66.27} &
  \multicolumn{1}{c|}{69.41} &
  \multicolumn{1}{c|}{67.84} &
  \multicolumn{1}{c|}{78.62} &
  \multicolumn{1}{c|}{80.74} &
   79.68 &
  \multicolumn{1}{c|}{53.77} &
  \multicolumn{1}{c|}{55.94} &
  \multicolumn{1}{c|}{54.96} &
  \multicolumn{1}{c|}{61.67} &
  \multicolumn{1}{c|}{67.02} &
  64.35 \\ \hline
ProtoNet + Ours &
  \multicolumn{1}{c|}{69.39} &
  \multicolumn{1}{c|}{72.33} &
  \multicolumn{1}{c|}{\textbf{70.84}} &
  \multicolumn{1}{c|}{74.07} &
  \multicolumn{1}{c|}{75.34} &
  \textbf{74.71} &
  \multicolumn{1}{c|}{48.98} &
  \multicolumn{1}{c|}{52.62} &
  \multicolumn{1}{c|}{\textbf{50.8}} &
  \multicolumn{1}{c|}{58.85} &
  \multicolumn{1}{c|}{64.26} &
  \multicolumn{1}{c}{\textbf{61.56}}
\\ \hline
AttMPTI + Ours &
  \multicolumn{1}{c|}{74.30} &
  \multicolumn{1}{c|}{77.62} &
  \multicolumn{1}{c|}{\textbf{75.96}} &
  \multicolumn{1}{c|}{81.86} &
  \multicolumn{1}{c|}{82.39} &
  \textbf{82.13} &
  \multicolumn{1}{c|}{58.85} &
  \multicolumn{1}{c|}{60.29} &
  \multicolumn{1}{c|}{\textbf{59.57}} &
  \multicolumn{1}{c|}{66.56} &
  \multicolumn{1}{c|}{79.46} &
  \multicolumn{1}{c}{\textbf{69.01}} \\ \hline
\multicolumn{13}{c}{ScanNet} \\ \hline
ProtoNet~\cite{prototype_cls} &
  \multicolumn{1}{c|}{55.98} &
  \multicolumn{1}{c|}{57.81} &
  \multicolumn{1}{c|}{56.90} &
  \multicolumn{1}{c|}{59.51} &
  \multicolumn{1}{c|}{63.46} &
  61.49 &
  \multicolumn{1}{c|}{30.95} &
  \multicolumn{1}{c|}{33.92} &
  \multicolumn{1}{c|}{32.44} &
  \multicolumn{1}{c|}{42.01} &
  \multicolumn{1}{c|}{45.34} &
  43.68 \\ \hline
AttMPTI \cite{zhao2021few} &
  \multicolumn{1}{c|}{56.67} &
  \multicolumn{1}{c|}{59.79} &
  \multicolumn{1}{c|}{58.23} &
  \multicolumn{1}{c|}{66.70} &
  \multicolumn{1}{c|}{70.29} &
  68.50 &
  \multicolumn{1}{c|}{40.83} &
  \multicolumn{1}{c|}{42.55} &
  \multicolumn{1}{c|}{41.69} &
  \multicolumn{1}{c|}{50.32} &
  \multicolumn{1}{c|}{54.00} &
  52.16 \\ \hline
ProtoNet + Ours &
  \multicolumn{1}{c|}{57.40} &
  \multicolumn{1}{c|}{59.31} &
  \multicolumn{1}{c|}{\textbf{58.36}} &
  \multicolumn{1}{c|}{60.83} &
  \multicolumn{1}{c|}{66.01} &
  \textbf{63.42} &
  \multicolumn{1}{c|}{37.18} &
  \multicolumn{1}{c|}{39.28} &
  \multicolumn{1}{c|}{\textbf{38.23}} &
  \multicolumn{1}{c|}{44.11} &
  \multicolumn{1}{c|}{47.01} &
  \multicolumn{1}{c}{\textbf{45.56}}
\\ \hline
AttMPTI + Ours &
  \multicolumn{1}{c|}{59.06} &
  \multicolumn{1}{c|}{61.66} &
  \multicolumn{1}{c|}{\textbf{60.36}} &
  \multicolumn{1}{c|}{66.88} &
  \multicolumn{1}{c|}{72.17} &
  \textbf{69.53} &
  \multicolumn{1}{c|}{43.10} &
  \multicolumn{1}{c|}{46.79} &
  \multicolumn{1}{c|}{\textbf{44.95}} &
  \multicolumn{1}{c|}{51.91} &
  \multicolumn{1}{c|}{57.21} &
  \multicolumn{1}{c}{\textbf{54.56}} \\
  \bottomrule
\end{tabular}
\caption{The results of ProtoNet and AttMPTI, as well as our proposed method, were evaluated on S3DIS and ScanNet datasets using the mean-IoU metric (\%). The notation $S^{i}$ represents the $i$-\textit{th} split  for testing.}
\label{tab:results_s3dis_scannet}
\vspace{-0.1cm}
\end{table*}

\begin{figure}[t]
\centering
\includegraphics[width=\linewidth]{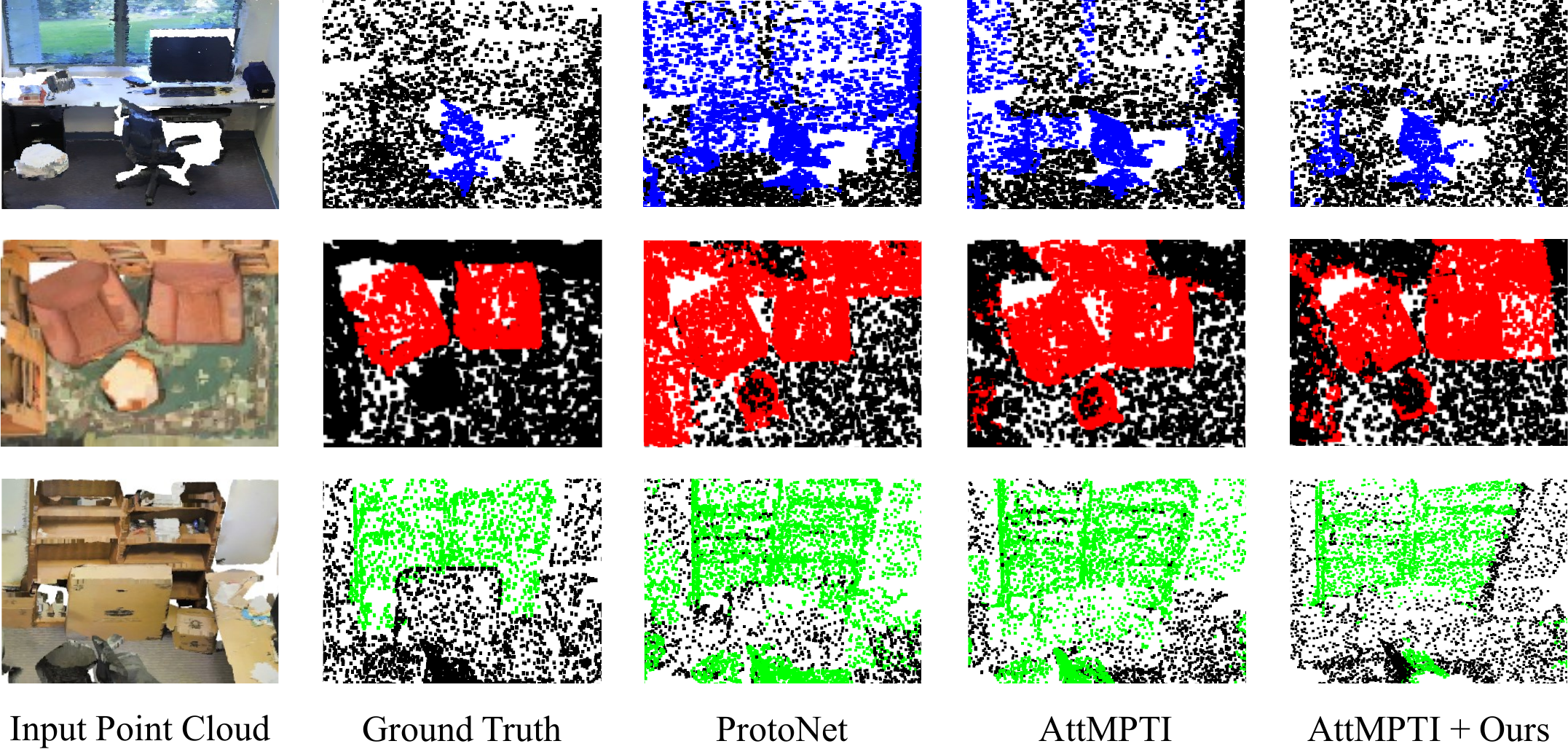}
\caption{Visual illustrations from left to right are input point cloud, ground truth, ProtoNet, AttMPTI and AttMPTI + Ours. The target classes from top to bottom are `chair' (blue), `sofa' (red) and `bookcase' (green).}
\label{fig:pred_visualization}
\vspace{0.1cm}
\end{figure}

\begin{table}[t]
\resizebox{\linewidth}{!}{
    \begin{tabular}{lcccc}
    \toprule
    Method & ProtoNet & ProtoNet+Ours  & AttMPTI  &   AttMPTI+Ours                 \\
    \midrule
    Inference time(ms)    & 12.0   & 12.7    & 133.2   & 133.5                  \\
    \bottomrule
    \end{tabular}
}
\caption{Efficiency comparison between the baselines and our method. The average inference time for a single forward is reported.}
\label{tab:inference_time_comparison}
\vspace{-0.5cm}
\end{table}

To validate the efficacy of our approach, we perform 1-way 1-shot, 1-way 5-shot, 2-way 1-shot, and 2-way 5-shot experiments on both S3DIS and ScanNet datasets. We employ ProtoNet and AttMPTI~\cite{zhao2021few} as our baselines and apply our method to them to manifest the generalization ability.

Table \ref{tab:results_s3dis_scannet} depicts the results obtained on the S3DIS and ScanNet datasets, respectively. Across various evaluation settings on both benchmarks, our approach consistently yields substantial improvements over both base models. Despite the increased complexity of the ScanNet dataset, which consists of numerous diverse scenes, our method still brings impressive performance gains. Besides, the comparison regarding efficiency is shown in Table~\ref{tab:inference_time_comparison}. These results underscore the potential for our method to serve as an effective module that can be widely utilized to boost performance for different models in point-cloud few-shot segmentation tasks without compromising efficiency. 

The results demonstrate that our method works better in improving AttMPTI's performance in most cases, compared to ProtoNet. We attribute this observation to the fact that the semantic information of the foreground and background prototypes adopted in ProtoNet, generated using global average pooling, may be too coarse to fully leverage the potential of our method. In contrast, AttMPTI exploits multiple prototypes to represent foreground and background, and our method benefits from this strategy to achieve better background adaptation and holistic rectification.

\mypara{Qualitative results.} To aid in understanding, we present t-SNE results in Figure \ref{fig:tsne1}, demonstrating that our method helps the model yield more discriminative representations.
Besides, prediction visualizations are shown in Figure \ref{fig:pred_visualization} where our proposed method shows more visually attractive results.

\subsection{Ablation Study}
\label{sec:ablation_study}
In this section, we perform an ablation study to investigate the efficacy of the individual components comprising our proposed method. We conduct experiments on $S^0$ of S3DIS under the 1-way 1-shot setting and the state-of-the-art AttMPTI~\cite{zhao2021few} is adopted as our baseline model.

\begin{table}[t]
\tabcolsep=0.9cm
\begin{tabular}{lcc}
\toprule
\multicolumn{2}{l}{Methods}                     & \multicolumn{1}{c}{mIoU(\%)}  \\
\midrule
\multicolumn{2}{l}{(a) AttMPTI}                 & 66.27                         \\
\multicolumn{2}{l}{(b) + BPA (BG)}              & 71.36                         \\
\multicolumn{2}{l}{(c) + BPA (BG + FG)}         & 63.39                         \\
\multicolumn{2}{l}{(d) + HR ($L_1$ Loss)}       & 67.91                         \\
\multicolumn{2}{l}{(e) + HR ($L_2$ Loss)}       & 68.16                         \\
\multicolumn{2}{l}{(f) + HR (KL)}               & 69.54                         \\
\multicolumn{2}{l}{(g) + BPA (BG) + HR (KL)}    & \textbf{74.30}                \\
\bottomrule
\end{tabular}
\caption{Ablation study on BPA and HR.}
\label{tab:ablation_study_component}
\vspace{-0.5cm}
\end{table}

\mypara{Effectiveness of individual components. } We investigate the impacts of ackground Prototype Adaptation (BPA) and Holistic Rectification (HR), and report the results in Table \ref{tab:ablation_study_component}.

Notably, both BPA and HR demonstrate significant improvements to the baseline in experiments (b) and (f). However, applying the same BPA strategy to the support foreground prototype $\vec{p}_s^{fg}$ in experiment (c) did not yield any further enhancements as discussed in Section~\ref{sec:bpa}.

With regards to HR, we compared our distillation-based approach, which utilizes KL divergence to align the predicted logits, against a direct application of the $L_1$/$L_2$ losses between $\vec{p}_s$ and $\vec{p}_q$. The results of experiments (d), (e) and (f) demonstrate that employing KL divergence yielded superior performance.

The proposed methods are ultimately integrated into the last experiment (g) presented in Table~\ref{tab:ablation_study_component}, where their complementary nature is demonstrated by the result.

\mypara{Effects of different loss weights. } In order to investigate the impact of varying values of $\lambda_{KL}$ in our distillation loss, we conducted experiments on S3DIS under the 1-way 1-shot $S^0$ setting. The results presented in Table \ref{tab:ablation_study_loss_weight} indicate that setting $\lambda_{KL}$ to $1$ produces satisfactory performance.

\begin{table}[t]
\tabcolsep=0.8cm
\begin{tabular}{lc}
\toprule
Training Objectives      & mIoU(\%)  \\
\midrule
$\mathcal{L} = \mathcal{L}_{CE} + \mathcal{L}_{CE}^{o} + 0.1 \cdot \mathcal{L}_{KL}$   & 73.01  \\
$\mathcal{L} = \mathcal{L}_{CE} + \mathcal{L}_{CE}^{o} + \mathcal{L}_{KL}$             & \textbf{74.30} \\
$\mathcal{L} = \mathcal{L}_{CE} + \mathcal{L}_{CE}^{o} + 10 \cdot \mathcal{L}_{KL}$    & 74.29  \\
$\mathcal{L} = \mathcal{L}_{CE} + \mathcal{L}_{CE}^{o} + 20 \cdot \mathcal{L}_{KL}$    & 68.79   \\
\bottomrule
\end{tabular}
\caption{The results of different values for $\lambda_{KL}$.}
\label{tab:ablation_study_loss_weight}
\vspace{-0.3cm}
\end{table}

\mypara{Components of BPA. } Our proposed BPA can be partitioned into three distinct components: 1) query information adaptation component, $i.e.$, $\mathcal{r}_q = \mathcal{C}_1(\vec{p}_s^{bg}, \vec{f}_q)$ in Eq.~\eqref{eq:query_adaptation}, 2) support foreground information filtering component, $i.e.$, Eq.~\eqref{eq:FG_information_filter}, and 3) adapted prototype rectification module, $i.e.$, $\vec{p}_{as}^{bg}=\mathcal{R}(\vec{p}_s^{bg}, \mathcal{r})$ in Eq.~\eqref{eq:proto_adapt}, each of which contains a cross-attention module.  Our experiments revealed that each component of our design plays a crucial role in improving performance, and results are presented in Table \ref{tab:prototype_adaptation_design}. 

To elaborate, the query information adaptation component of BPA takes $\vec{p}_s^{bg}$ as the query ($q$) and $\vec{f}_q$ as the key-value ($k,v$) in cross-attention. For the results without check on $\mathcal{r}_q = \mathcal{C}_1(\vec{p}_s^{bg}, \vec{f}_q)$, it will instead be $\mathcal{r}_q=\vec{p_s^{bg}}$. The support foreground information filtering component in Eq.~\eqref{eq:FG_information_filter}, on the other hand, utilizes $\vec{p}_s^{bg}$ as the query ($q$) and $\vec{p}_s^{fg}$ as the key-value ($k,v$) in cross-attention.  For models without the checks on $\mathcal{G}(\mathcal{r}_q,  \mathcal{r}_s)$, there will be a direct assignment as $\mathcal{r}=\mathcal{r}_q$. Finally, the support background prototype projection with $\mathcal{R}$ employs $\vec{p}_s^{bg}$ as the query ($q$) and $\vec{p}_d^{bg}$ as the key-value ($k,v$) in cross-attention. In this case, the results without checks are obtained with $\vec{p}_{as}^{bg}=\mathcal{r}$, which means the output is not conditioned on $\vec{p}_s^{bg}$. Based on the comparisons shown in Table~\ref{tab:prototype_adaptation_design}, we can see that both components are indispensable.

\begin{table}[t]
\tabcolsep=0.3cm
\begin{tabular}{l|ccccc}
\toprule             
$\mathcal{C}_1(\vec{p}_s^{bg}, \vec{f}_q)$       &                &               & $\checkmark$  & $\checkmark$   & $\checkmark$     \\
$\mathcal{G}(\mathcal{r}_q,  \mathcal{r}_s)$     &                & $\checkmark$  &               & $\checkmark$   & $\checkmark$     \\
$\mathcal{R}(\vec{p}_s^{bg}, \mathcal{r})$       &                & $\checkmark$  & $\checkmark$  &                & $\checkmark$     \\
\midrule
mIoU(\%)                    & 66.27          & 68.83         & 61.22         & 69.36          & \textbf{71.36}   \\
\bottomrule
\end{tabular}
\caption{The experiments of individual components of BPA.}
\label{tab:prototype_adaptation_design}
\vspace{-0.5cm}
\end{table}

\section{Related Works}
\label{sec:related_work}

\mypara{Few-shot learning. }
Few-shot learning is a challenging task that involves making accurate predictions on new classes with only a limited number of labeled examples. Popular solutions to this problem include meta-learning based methods \cite{memory_match,maml,reslt,gpaco,leo}, as well as metric-learning-based approaches such as MatchingNet \cite{matchingnet}, RelationNet \cite{relationnet}, ProtoNet \cite{prototype_cls}, and DeepEMD \cite{deepemd}. Furthermore, data augmentation techniques have been shown to improve model performance by mitigating overfitting \cite{hallucinate_saliency,hallucinating,imaginary}. The frameworks adopted in this work, AttMPTI \cite{zhao2021few}, are both built upon ProtoNet \cite{prototype_cls}, which belongs to the metric-learning family.

\mypara{Point cloud segmentation. }
Point cloud segmentation is a challenging task that involves classifying each point in a point cloud into known classes. In recent years, point-based approaches have emerged as the mainstream method for addressing this task, including PointNet \cite{qi2017pointnet}, PointNet++ \cite{NIPS2017_d8bf84be}, DGCNN \cite{wang2019dynamic}, KPConv \cite{thomas2019kpconv}, RandLA-Net \cite{hu2020randla}, and Point Transformer V1 \& V2 \cite{zhao2021point,wu2022point}, among others.
DGCNN \cite{wang2019dynamic} introduced the EdgeConv module to capture local structures in point clouds and utilized a dynamic $k$-NN graph to achieve long-distance point intersections. AttMPTI~\cite{zhao2021few} adopt DGCNN as the feature extractor.

\mypara{Few-shot segmentation. }
Few-shot point cloud segmentation (PC-FSS) combines the challenges of point cloud segmentation with the limited availability of annotated data. While there are numerous works on few-shot image segmentation \cite{shaban,pfenetpp,canet,pfenet,hdmnet,tian2022gfsseg,repri,ppnet}, PC-FSS remains an under-explored research area. The idea of graph reasoning~\cite{chen2023generalizing,liang2022reasoning,luo2023normalizing,zhao2023towards,liang2023message,liang2023structure,luo2023normalizing} is adopted in FSS and found effective. Motivated by the graph propagation, 
AttMPTI \cite{zhao2021few} is a representative work in PC-FSS, proposing an attention-aware multi-prototypes transductive inference method that has shown promise in PC-FSS.

\section{Concluding Remarks}
\label{ref:conclusion}
We have introduced our proposed method for PC-FSS, which incorporates two novel query-guided enhancement strategies: Background Prototype Adaptation (BPA) and Holistic Rectification (HR). BPA enables the adaptive adjustment of the content of support background prototypes to better align with the individual query background context, resulting in more representative background prototypes. In contrast, since BPA cannot be effectively applied to the support foreground prototypes, HR performs holistic rectification on support prototypes by encouraging them to behave like their optimal counterparts. Both contributions are found to be beneficial, with minimal additional costs added to the baselines. We hope that our designs can serve as plug-in modules for various baselines and benefit future research in this field.


\bibliographystyle{ACM-Reference-Format}
\balance
\bibliography{base}


\begin{thebibliography}{59}

%%% the Web address.

\ifx \showCODEN    \undefined \def \showCODEN     #1{\unskip}     \fi
\ifx \showDOI      \undefined \def \showDOI       #1{#1}\fi
\ifx \showISBNx    \undefined \def \showISBNx     #1{\unskip}     \fi
\ifx \showISBNxiii \undefined \def \showISBNxiii  #1{\unskip}     \fi
\ifx \showISSN     \undefined \def \showISSN      #1{\unskip}     \fi
\ifx \showLCCN     \undefined \def \showLCCN      #1{\unskip}     \fi
\ifx \shownote     \undefined \def \shownote      #1{#1}          \fi
\ifx \showarticletitle \undefined \def \showarticletitle #1{#1}   \fi
\ifx \showURL      \undefined \def \showURL       {\relax}        \fi
% invisible to TeX
\providecommand\bibfield[2]{#2}
\providecommand\bibinfo[2]{#2}
\providecommand\natexlab[1]{#1}
\providecommand\showeprint[2][]{arXiv:#2}

\bibitem[Ao et~al\mbox{.}(2021)]%
        {ao2020SpinNet}
\bibfield{author}{\bibinfo{person}{Sheng Ao}, \bibinfo{person}{Qingyong Hu},
  \bibinfo{person}{Bo Yang}, \bibinfo{person}{Andrew Markham}, {and}
  \bibinfo{person}{Yulan Guo}.} \bibinfo{year}{2021}\natexlab{}.
\newblock \showarticletitle{SpinNet: Learning a General Surface Descriptor for
  3D Point Cloud Registration}. In \bibinfo{booktitle}{\emph{CVPR}}.
\newblock


\bibitem[{Armeni} et~al\mbox{.}(2017)]%
        {2017arXiv170201105A}
\bibfield{author}{\bibinfo{person}{I. {Armeni}}, \bibinfo{person}{A. {Sax}},
  \bibinfo{person}{A.~R. {Zamir}}, {and} \bibinfo{person}{S. {Savarese}}.}
  \bibinfo{year}{2017}\natexlab{}.
\newblock \showarticletitle{Joint 2D-3D-Semantic Data for Indoor Scene
  Understanding}.
\newblock \bibinfo{journal}{\emph{ArXiv}} (\bibinfo{year}{2017}).
\newblock


\bibitem[Boudiaf et~al\mbox{.}(2021)]%
        {repri}
\bibfield{author}{\bibinfo{person}{Malik Boudiaf}, \bibinfo{person}{Hoel
  Kervadec}, \bibinfo{person}{Imtiaz~Masud Ziko}, \bibinfo{person}{Pablo
  Piantanida}, \bibinfo{person}{Ismail~Ben Ayed}, {and} \bibinfo{person}{Jose
  Dolz}.} \bibinfo{year}{2021}\natexlab{}.
\newblock \showarticletitle{Few-Shot Segmentation Without Meta-Learning: {A}
  Good Transductive Inference Is All You Need?}. In
  \bibinfo{booktitle}{\emph{CVPR}}.
\newblock


\bibitem[Cai et~al\mbox{.}(2018)]%
        {memory_match}
\bibfield{author}{\bibinfo{person}{Qi Cai}, \bibinfo{person}{Yingwei Pan},
  \bibinfo{person}{Ting Yao}, \bibinfo{person}{Chenggang Yan}, {and}
  \bibinfo{person}{Tao Mei}.} \bibinfo{year}{2018}\natexlab{}.
\newblock \showarticletitle{Memory Matching Networks for One-Shot Image
  Recognition}. In \bibinfo{booktitle}{\emph{CVPR}}.
\newblock


\bibitem[Chen et~al\mbox{.}(2023)]%
        {chen2023generalizing}
\bibfield{author}{\bibinfo{person}{Mingyang Chen}, \bibinfo{person}{Wen Zhang},
  \bibinfo{person}{Yuxia Geng}, \bibinfo{person}{Zezhong Xu},
  \bibinfo{person}{Jeff~Z Pan}, {and} \bibinfo{person}{Huajun Chen}.}
  \bibinfo{year}{2023}\natexlab{}.
\newblock \showarticletitle{Generalizing to Unseen Elements: A Survey on
  Knowledge Extrapolation for Knowledge Graphs}.
\newblock \bibinfo{journal}{\emph{arXiv preprint arXiv:2302.01859}}
  (\bibinfo{year}{2023}).
\newblock


\bibitem[Choy et~al\mbox{.}(2019)]%
        {choy20194d}
\bibfield{author}{\bibinfo{person}{Christopher Choy}, \bibinfo{person}{JunYoung
  Gwak}, {and} \bibinfo{person}{Silvio Savarese}.}
  \bibinfo{year}{2019}\natexlab{}.
\newblock \showarticletitle{4D Spatio-Temporal ConvNets: Minkowski
  Convolutional Neural Networks}. In \bibinfo{booktitle}{\emph{CVPR}}.
\newblock


\bibitem[Cui et~al\mbox{.}(2023a)]%
        {reslt}
\bibfield{author}{\bibinfo{person}{Jiequan Cui}, \bibinfo{person}{Shu Liu},
  \bibinfo{person}{Zhuotao Tian}, \bibinfo{person}{Zhisheng Zhong}, {and}
  \bibinfo{person}{Jiaya Jia}.} \bibinfo{year}{2023}\natexlab{a}.
\newblock \showarticletitle{ResLT: Residual Learning for Long-Tailed
  Recognition}.
\newblock \bibinfo{journal}{\emph{TPAMI}} \bibinfo{volume}{45},
  \bibinfo{number}{3} (\bibinfo{year}{2023}), \bibinfo{pages}{3695--3706}.
\newblock


\bibitem[Cui et~al\mbox{.}(2023b)]%
        {decouplekld}
\bibfield{author}{\bibinfo{person}{Jiequan Cui}, \bibinfo{person}{Zhuotao
  Tian}, \bibinfo{person}{Zhisheng Zhong}, \bibinfo{person}{Xiaojuan Qi},
  \bibinfo{person}{Bei Yu}, {and} \bibinfo{person}{Hanwang Zhang}.}
  \bibinfo{year}{2023}\natexlab{b}.
\newblock \showarticletitle{Decoupled Kullback-Leibler Divergence Loss}.
\newblock \bibinfo{journal}{\emph{CoRR}}  \bibinfo{volume}{abs/2305.13948}
  (\bibinfo{year}{2023}).
\newblock
\urldef\tempurl%
\url{https://doi.org/10.48550/arXiv.2305.13948}
\showDOI{\tempurl}
\showeprint[arXiv]{2305.13948}


\bibitem[Cui et~al\mbox{.}(2022a)]%
        {rr}
\bibfield{author}{\bibinfo{person}{Jiequan Cui}, \bibinfo{person}{Yuhui Yuan},
  \bibinfo{person}{Zhisheng Zhong}, \bibinfo{person}{Zhuotao Tian},
  \bibinfo{person}{Han Hu}, \bibinfo{person}{Stephen Lin}, {and}
  \bibinfo{person}{Jiaya Jia}.} \bibinfo{year}{2022}\natexlab{a}.
\newblock \showarticletitle{Region Rebalance for Long-Tailed Semantic
  Segmentation}.
\newblock \bibinfo{journal}{\emph{CoRR}}  \bibinfo{volume}{abs/2204.01969}
  (\bibinfo{year}{2022}).
\newblock
\urldef\tempurl%
\url{https://doi.org/10.48550/arXiv.2204.01969}
\showDOI{\tempurl}
\showeprint[arXiv]{2204.01969}


\bibitem[Cui et~al\mbox{.}(2022b)]%
        {gpaco}
\bibfield{author}{\bibinfo{person}{Jiequan Cui}, \bibinfo{person}{Zhisheng
  Zhong}, \bibinfo{person}{Zhuotao Tian}, \bibinfo{person}{Shu Liu},
  \bibinfo{person}{Bei Yu}, {and} \bibinfo{person}{Jiaya Jia}.}
  \bibinfo{year}{2022}\natexlab{b}.
\newblock \showarticletitle{Generalized Parametric Contrastive Learning}.
\newblock \bibinfo{journal}{\emph{CoRR}}  \bibinfo{volume}{abs/2209.12400}
  (\bibinfo{year}{2022}).
\newblock
\urldef\tempurl%
\url{https://doi.org/10.48550/arXiv.2209.12400}
\showDOI{\tempurl}
\showeprint[arXiv]{2209.12400}


\bibitem[Dai et~al\mbox{.}(2017)]%
        {dai2017scannet}
\bibfield{author}{\bibinfo{person}{Angela Dai}, \bibinfo{person}{Angel~X.
  Chang}, \bibinfo{person}{Manolis Savva}, \bibinfo{person}{Maciej Halber},
  \bibinfo{person}{Thomas Funkhouser}, {and} \bibinfo{person}{Matthias
  Nie{\ss}ner}.} \bibinfo{year}{2017}\natexlab{}.
\newblock \showarticletitle{ScanNet: Richly-annotated 3D Reconstructions of
  Indoor Scenes}. In \bibinfo{booktitle}{\emph{CVPR}}.
\newblock


\bibitem[Deng et~al\mbox{.}(2020)]%
        {deng2020voxel}
\bibfield{author}{\bibinfo{person}{Jiajun Deng}, \bibinfo{person}{Shaoshuai
  Shi}, \bibinfo{person}{Peiwei Li}, \bibinfo{person}{Wengang Zhou},
  \bibinfo{person}{Yanyong Zhang}, {and} \bibinfo{person}{Houqiang Li}.}
  \bibinfo{year}{2020}\natexlab{}.
\newblock \showarticletitle{Voxel R-CNN: Towards High Performance Voxel-based
  3D Object Detection}.
\newblock \bibinfo{journal}{\emph{arXiv}} (\bibinfo{year}{2020}).
\newblock


\bibitem[Finn et~al\mbox{.}(2017)]%
        {maml}
\bibfield{author}{\bibinfo{person}{Chelsea Finn}, \bibinfo{person}{Pieter
  Abbeel}, {and} \bibinfo{person}{Sergey Levine}.}
  \bibinfo{year}{2017}\natexlab{}.
\newblock \showarticletitle{Model-Agnostic Meta-Learning for Fast Adaptation of
  Deep Networks}. In \bibinfo{booktitle}{\emph{ICML}}.
\newblock


\bibitem[Hariharan and Girshick(2017)]%
        {hallucinating}
\bibfield{author}{\bibinfo{person}{Bharath Hariharan} {and}
  \bibinfo{person}{Ross~B. Girshick}.} \bibinfo{year}{2017}\natexlab{}.
\newblock \showarticletitle{Low-Shot Visual Recognition by Shrinking and
  Hallucinating Features}. In \bibinfo{booktitle}{\emph{ICCV}}.
\newblock


\bibitem[Hinton et~al\mbox{.}(2015)]%
        {hinton2015distilling}
\bibfield{author}{\bibinfo{person}{Geoffrey Hinton}, \bibinfo{person}{Oriol
  Vinyals}, {and} \bibinfo{person}{Jeff Dean}.}
  \bibinfo{year}{2015}\natexlab{}.
\newblock \showarticletitle{Distilling the knowledge in a neural network}.
\newblock \bibinfo{journal}{\emph{ArXiv}} (\bibinfo{year}{2015}).
\newblock


\bibitem[Hu et~al\mbox{.}(2020)]%
        {hu2020randla}
\bibfield{author}{\bibinfo{person}{Qingyong Hu}, \bibinfo{person}{Bo Yang},
  \bibinfo{person}{Linhai Xie}, \bibinfo{person}{Stefano Rosa},
  \bibinfo{person}{Yulan Guo}, \bibinfo{person}{Zhihua Wang},
  \bibinfo{person}{Niki Trigoni}, {and} \bibinfo{person}{Andrew Markham}.}
  \bibinfo{year}{2020}\natexlab{}.
\newblock \showarticletitle{Randla-net: Efficient semantic segmentation of
  large-scale point clouds}. In \bibinfo{booktitle}{\emph{CVPR}}.
\newblock


\bibitem[Iscen et~al\mbox{.}(2019)]%
        {iscen2019label}
\bibfield{author}{\bibinfo{person}{Ahmet Iscen}, \bibinfo{person}{Giorgos
  Tolias}, \bibinfo{person}{Yannis Avrithis}, {and} \bibinfo{person}{Ondrej
  Chum}.} \bibinfo{year}{2019}\natexlab{}.
\newblock \showarticletitle{Label propagation for deep semi-supervised
  learning}. In \bibinfo{booktitle}{\emph{CVPR}}.
\newblock


\bibitem[Jiang et~al\mbox{.}(2021a)]%
        {3dsemiseg}
\bibfield{author}{\bibinfo{person}{Li Jiang}, \bibinfo{person}{Shaoshuai Shi},
  \bibinfo{person}{Zhuotao Tian}, \bibinfo{person}{Xin Lai},
  \bibinfo{person}{Shu Liu}, \bibinfo{person}{Chi{-}Wing Fu}, {and}
  \bibinfo{person}{Jiaya Jia}.} \bibinfo{year}{2021}\natexlab{a}.
\newblock \showarticletitle{Guided Point Contrastive Learning for
  Semi-supervised Point Cloud Semantic Segmentation}. In
  \bibinfo{booktitle}{\emph{ICCV}}.
\newblock


\bibitem[Jiang et~al\mbox{.}(2021b)]%
        {jiang2021guided}
\bibfield{author}{\bibinfo{person}{Li Jiang}, \bibinfo{person}{Shaoshuai Shi},
  \bibinfo{person}{Zhuotao Tian}, \bibinfo{person}{Xin Lai},
  \bibinfo{person}{Shu Liu}, \bibinfo{person}{Chi-Wing Fu}, {and}
  \bibinfo{person}{Jiaya Jia}.} \bibinfo{year}{2021}\natexlab{b}.
\newblock \showarticletitle{Guided Point Contrastive Learning for
  Semi-supervised Point Cloud Semantic Segmentation}. In
  \bibinfo{booktitle}{\emph{ICCV}}.
\newblock


\bibitem[Lai et~al\mbox{.}(2022a)]%
        {stratified}
\bibfield{author}{\bibinfo{person}{Xin Lai}, \bibinfo{person}{Jianhui Liu},
  \bibinfo{person}{Li Jiang}, \bibinfo{person}{Liwei Wang},
  \bibinfo{person}{Hengshuang Zhao}, \bibinfo{person}{Shu Liu},
  \bibinfo{person}{Xiaojuan Qi}, {and} \bibinfo{person}{Jiaya Jia}.}
  \bibinfo{year}{2022}\natexlab{a}.
\newblock \showarticletitle{Stratified Transformer for 3D Point Cloud
  Segmentation}. In \bibinfo{booktitle}{\emph{CVPR}}.
\newblock


\bibitem[Lai et~al\mbox{.}(2021)]%
        {semiseg}
\bibfield{author}{\bibinfo{person}{Xin Lai}, \bibinfo{person}{Zhuotao Tian},
  \bibinfo{person}{Li Jiang}, \bibinfo{person}{Shu Liu},
  \bibinfo{person}{Hengshuang Zhao}, \bibinfo{person}{Liwei Wang}, {and}
  \bibinfo{person}{Jiaya Jia}.} \bibinfo{year}{2021}\natexlab{}.
\newblock \showarticletitle{Semi-Supervised Semantic Segmentation With
  Directional Context-Aware Consistency}. In \bibinfo{booktitle}{\emph{CVPR}}.
\newblock


\bibitem[Lai et~al\mbox{.}(2022b)]%
        {decouplenet}
\bibfield{author}{\bibinfo{person}{Xin Lai}, \bibinfo{person}{Zhuotao Tian},
  \bibinfo{person}{Xiaogang Xu}, \bibinfo{person}{Ying{-}Cong Chen},
  \bibinfo{person}{Shu Liu}, \bibinfo{person}{Hengshuang Zhao},
  \bibinfo{person}{Liwei Wang}, {and} \bibinfo{person}{Jiaya Jia}.}
  \bibinfo{year}{2022}\natexlab{b}.
\newblock \showarticletitle{DecoupleNet: Decoupled Network for Domain Adaptive
  Semantic Segmentation}. In \bibinfo{booktitle}{\emph{ECCV}},
  \bibfield{editor}{\bibinfo{person}{Shai Avidan}, \bibinfo{person}{Gabriel~J.
  Brostow}, \bibinfo{person}{Moustapha Ciss{\'{e}}},
  \bibinfo{person}{Giovanni~Maria Farinella}, {and} \bibinfo{person}{Tal
  Hassner}} (Eds.).
\newblock


\bibitem[Liang et~al\mbox{.}(2023a)]%
        {liang2023knowledge}
\bibfield{author}{\bibinfo{person}{Ke Liang}, \bibinfo{person}{Yue Liu},
  \bibinfo{person}{Sihang Zhou}, \bibinfo{person}{Wenxuan Tu},
  \bibinfo{person}{Yi Wen}, \bibinfo{person}{Xihong Yang},
  \bibinfo{person}{Xiangjun Dong}, {and} \bibinfo{person}{Xinwang Liu}.}
  \bibinfo{year}{2023}\natexlab{a}.
\newblock \showarticletitle{Knowledge Graph Contrastive Learning Based on
  Relation-Symmetrical Structure}.
\newblock \bibinfo{journal}{\emph{IEEE Transactions on Knowledge and Data
  Engineering}} (\bibinfo{year}{2023}).
\newblock


\bibitem[Liang et~al\mbox{.}(2022)]%
        {liang2022reasoning}
\bibfield{author}{\bibinfo{person}{Ke Liang}, \bibinfo{person}{Lingyuan Meng},
  \bibinfo{person}{Meng Liu}, \bibinfo{person}{Yue Liu},
  \bibinfo{person}{Wenxuan Tu}, \bibinfo{person}{Siwei Wang},
  \bibinfo{person}{Sihang Zhou}, \bibinfo{person}{Xinwang Liu}, {and}
  \bibinfo{person}{Fuchun Sun}.} \bibinfo{year}{2022}\natexlab{}.
\newblock \showarticletitle{Reasoning over Different Types of Knowledge Graphs:
  Static, Temporal and Multi-Modal}.
\newblock \bibinfo{journal}{\emph{arXiv preprint arXiv:2212.05767}}
  (\bibinfo{year}{2022}).
\newblock


\bibitem[Liang et~al\mbox{.}(2023b)]%
        {liang2023message}
\bibfield{author}{\bibinfo{person}{Ke Liang}, \bibinfo{person}{Lingyuan Meng},
  \bibinfo{person}{Sihang Zhou}, \bibinfo{person}{Siwei Wang},
  \bibinfo{person}{Wenxuan Tu}, \bibinfo{person}{Yue Liu},
  \bibinfo{person}{Meng Liu}, {and} \bibinfo{person}{Xinwang Liu}.}
  \bibinfo{year}{2023}\natexlab{b}.
\newblock \showarticletitle{Message Intercommunication for Inductive Relation
  Reasoning}.
\newblock \bibinfo{journal}{\emph{arXiv preprint arXiv:2305.14074}}
  (\bibinfo{year}{2023}).
\newblock


\bibitem[Liang et~al\mbox{.}(2023c)]%
        {liang2023structure}
\bibfield{author}{\bibinfo{person}{Ke Liang}, \bibinfo{person}{Sihang Zhou},
  \bibinfo{person}{Yue Liu}, \bibinfo{person}{Lingyuan Meng},
  \bibinfo{person}{Meng Liu}, {and} \bibinfo{person}{Xinwang Liu}.}
  \bibinfo{year}{2023}\natexlab{c}.
\newblock \showarticletitle{Structure Guided Multi-modal Pre-trained
  Transformer for Knowledge Graph Reasoning}.
\newblock \bibinfo{journal}{\emph{arXiv preprint arXiv:2307.03591}}
  (\bibinfo{year}{2023}).
\newblock


\bibitem[Liu et~al\mbox{.}(2022)]%
        {efficient3d}
\bibfield{author}{\bibinfo{person}{Jianhui Liu}, \bibinfo{person}{Yukang Chen},
  \bibinfo{person}{Xiaoqing Ye}, \bibinfo{person}{Zhuotao Tian},
  \bibinfo{person}{Xiao Tan}, {and} \bibinfo{person}{Xiaojuan Qi}.}
  \bibinfo{year}{2022}\natexlab{}.
\newblock \showarticletitle{Spatial Pruned Sparse Convolution for Efficient 3D
  Object Detection}. In \bibinfo{booktitle}{\emph{NeurIPS}}.
\newblock


\bibitem[Liu(2016)]%
        {robot}
\bibfield{author}{\bibinfo{person}{Ming Liu}.} \bibinfo{year}{2016}\natexlab{}.
\newblock \showarticletitle{Robotic Online Path Planning on Point Cloud}.
\newblock \bibinfo{journal}{\emph{{IEEE} Trans. Cybern.}}
  (\bibinfo{year}{2016}).
\newblock


\bibitem[Liu et~al\mbox{.}(2020)]%
        {ppnet}
\bibfield{author}{\bibinfo{person}{Yongfei Liu}, \bibinfo{person}{Xiangyi
  Zhang}, \bibinfo{person}{Songyang Zhang}, {and} \bibinfo{person}{Xuming He}.}
  \bibinfo{year}{2020}\natexlab{}.
\newblock \showarticletitle{Part-aware Prototype Network for Few-shot Semantic
  Segmentation}. In \bibinfo{booktitle}{\emph{ECCV}}.
\newblock


\bibitem[Luo et~al\mbox{.}(2023)]%
        {luo2023normalizing}
\bibfield{author}{\bibinfo{person}{Linhao Luo}, \bibinfo{person}{Yuan-Fang Li},
  \bibinfo{person}{Gholamreza Haffari}, {and} \bibinfo{person}{Shirui Pan}.}
  \bibinfo{year}{2023}\natexlab{}.
\newblock \showarticletitle{Normalizing flow-based neural process for few-shot
  knowledge graph completion}. In \bibinfo{booktitle}{\emph{The 46th
  International ACM SIGIR Conference on Research and Development in Information
  Retrieval}}.
\newblock


\bibitem[Luo et~al\mbox{.}(2021)]%
        {pfenetpp}
\bibfield{author}{\bibinfo{person}{Xiaoliu Luo}, \bibinfo{person}{Zhuotao
  Tian}, \bibinfo{person}{Taiping Zhang}, \bibinfo{person}{Bei Yu},
  \bibinfo{person}{Yuan~Yan Tang}, {and} \bibinfo{person}{Jiaya Jia}.}
  \bibinfo{year}{2021}\natexlab{}.
\newblock \showarticletitle{PFENet++: Boosting Few-shot Semantic Segmentation
  with the Noise-filtered Context-aware Prior Mask}.
\newblock \bibinfo{journal}{\emph{CoRR}}  \bibinfo{volume}{abs/2109.13788}
  (\bibinfo{year}{2021}).
\newblock
\showeprint[arXiv]{2109.13788}
\urldef\tempurl%
\url{https://arxiv.org/abs/2109.13788}
\showURL{%
\tempurl}


\bibitem[Pan et~al\mbox{.}(2023)]%
        {llm_kg}
\bibfield{author}{\bibinfo{person}{Shirui Pan}, \bibinfo{person}{Linhao Luo},
  \bibinfo{person}{Yufei Wang}, \bibinfo{person}{Chen Chen},
  \bibinfo{person}{Jiapu Wang}, {and} \bibinfo{person}{Xindong Wu}.}
  \bibinfo{year}{2023}\natexlab{}.
\newblock \showarticletitle{Unifying Large Language Models and Knowledge
  Graphs: A Roadmap}.
\newblock \bibinfo{journal}{\emph{arXiv preprint arxiv:306.08302}}
  (\bibinfo{year}{2023}).
\newblock


\bibitem[Peng et~al\mbox{.}(2023)]%
        {hdmnet}
\bibfield{author}{\bibinfo{person}{Bohao Peng}, \bibinfo{person}{Zhuotao Tian},
  \bibinfo{person}{Xiaoyang Wu}, \bibinfo{person}{Chenyao Wang},
  \bibinfo{person}{Shu Liu}, \bibinfo{person}{Jingyong Su}, {and}
  \bibinfo{person}{Jiaya Jia}.} \bibinfo{year}{2023}\natexlab{}.
\newblock \showarticletitle{Hierarchical Dense Correlation Distillation for
  Few-Shot Segmentation}.
\newblock \bibinfo{journal}{\emph{CoRR}}  \bibinfo{volume}{abs/2303.14652}
  (\bibinfo{year}{2023}).
\newblock
\urldef\tempurl%
\url{https://doi.org/10.48550/arXiv.2303.14652}
\showDOI{\tempurl}
\showeprint[arXiv]{2303.14652}


\bibitem[Qi et~al\mbox{.}(2017a)]%
        {qi2017pointnet}
\bibfield{author}{\bibinfo{person}{Charles~R Qi}, \bibinfo{person}{Hao Su},
  \bibinfo{person}{Kaichun Mo}, {and} \bibinfo{person}{Leonidas~J Guibas}.}
  \bibinfo{year}{2017}\natexlab{a}.
\newblock \showarticletitle{Pointnet: Deep learning on point sets for 3d
  classification and segmentation}. In \bibinfo{booktitle}{\emph{CVPR}}.
\newblock


\bibitem[Qi et~al\mbox{.}(2017b)]%
        {NIPS2017_d8bf84be}
\bibfield{author}{\bibinfo{person}{Charles~Ruizhongtai Qi}, \bibinfo{person}{Li
  Yi}, \bibinfo{person}{Hao Su}, {and} \bibinfo{person}{Leonidas~J Guibas}.}
  \bibinfo{year}{2017}\natexlab{b}.
\newblock \showarticletitle{PointNet++: Deep Hierarchical Feature Learning on
  Point Sets in a Metric Space}. In \bibinfo{booktitle}{\emph{NeurIPS}}.
\newblock


\bibitem[Rusu et~al\mbox{.}(2019)]%
        {leo}
\bibfield{author}{\bibinfo{person}{Andrei~A. Rusu}, \bibinfo{person}{Dushyant
  Rao}, \bibinfo{person}{Jakub Sygnowski}, \bibinfo{person}{Oriol Vinyals},
  \bibinfo{person}{Razvan Pascanu}, \bibinfo{person}{Simon Osindero}, {and}
  \bibinfo{person}{Raia Hadsell}.} \bibinfo{year}{2019}\natexlab{}.
\newblock \showarticletitle{Meta-Learning with Latent Embedding Optimization}.
  In \bibinfo{booktitle}{\emph{ICLR}}.
\newblock


\bibitem[Shaban et~al\mbox{.}(2017)]%
        {shaban}
\bibfield{author}{\bibinfo{person}{Amirreza Shaban}, \bibinfo{person}{Shray
  Bansal}, \bibinfo{person}{Zhen Liu}, \bibinfo{person}{Irfan Essa}, {and}
  \bibinfo{person}{Byron Boots}.} \bibinfo{year}{2017}\natexlab{}.
\newblock \showarticletitle{One-Shot Learning for Semantic Segmentation}. In
  \bibinfo{booktitle}{\emph{BMVC}}.
\newblock


\bibitem[Snell et~al\mbox{.}(2017)]%
        {prototype_cls}
\bibfield{author}{\bibinfo{person}{Jake Snell}, \bibinfo{person}{Kevin
  Swersky}, {and} \bibinfo{person}{Richard~S. Zemel}.}
  \bibinfo{year}{2017}\natexlab{}.
\newblock \showarticletitle{Prototypical Networks for Few-shot Learning}. In
  \bibinfo{booktitle}{\emph{NeurIPS}}.
\newblock


\bibitem[Sung et~al\mbox{.}(2018)]%
        {relationnet}
\bibfield{author}{\bibinfo{person}{Flood Sung}, \bibinfo{person}{Yongxin Yang},
  \bibinfo{person}{Li Zhang}, \bibinfo{person}{Tao Xiang},
  \bibinfo{person}{Philip H.~S. Torr}, {and} \bibinfo{person}{Timothy~M.
  Hospedales}.} \bibinfo{year}{2018}\natexlab{}.
\newblock \showarticletitle{Learning to Compare: Relation Network for Few-Shot
  Learning}. In \bibinfo{booktitle}{\emph{CVPR}}.
\newblock


\bibitem[Thomas et~al\mbox{.}(2019)]%
        {thomas2019kpconv}
\bibfield{author}{\bibinfo{person}{Hugues Thomas}, \bibinfo{person}{Charles~R
  Qi}, \bibinfo{person}{Jean-Emmanuel Deschaud}, \bibinfo{person}{Beatriz
  Marcotegui}, \bibinfo{person}{Fran{\c{c}}ois Goulette}, {and}
  \bibinfo{person}{Leonidas~J Guibas}.} \bibinfo{year}{2019}\natexlab{}.
\newblock \showarticletitle{Kpconv: Flexible and deformable convolution for
  point clouds}. In \bibinfo{booktitle}{\emph{ICCV}}.
\newblock


\bibitem[Tian et~al\mbox{.}(2023a)]%
        {apd}
\bibfield{author}{\bibinfo{person}{Zhuotao Tian}, \bibinfo{person}{Pengguang
  Chen}, \bibinfo{person}{Xin Lai}, \bibinfo{person}{Li Jiang},
  \bibinfo{person}{Shu Liu}, \bibinfo{person}{Hengshuang Zhao},
  \bibinfo{person}{Bei Yu}, \bibinfo{person}{Ming{-}Chang Yang}, {and}
  \bibinfo{person}{Jiaya Jia}.} \bibinfo{year}{2023}\natexlab{a}.
\newblock \showarticletitle{Adaptive Perspective Distillation for Semantic
  Segmentation}.
\newblock \bibinfo{journal}{\emph{TPAMI}} \bibinfo{volume}{45},
  \bibinfo{number}{2} (\bibinfo{year}{2023}), \bibinfo{pages}{1372--1387}.
\newblock


\bibitem[Tian et~al\mbox{.}(2023b)]%
        {tian2023cac}
\bibfield{author}{\bibinfo{person}{Zhuotao Tian}, \bibinfo{person}{Jiequan
  Cui}, \bibinfo{person}{Li Jiang}, \bibinfo{person}{Xiaojuan Qi},
  \bibinfo{person}{Xin Lai}, \bibinfo{person}{Yixin Chen}, \bibinfo{person}{Shu
  Liu}, {and} \bibinfo{person}{Jiaya Jia}.} \bibinfo{year}{2023}\natexlab{b}.
\newblock \showarticletitle{Learning Context-aware Classifier for Semantic
  Segmentation}. In \bibinfo{booktitle}{\emph{Proceedings of the Thirty-Seventh
  {AAAI} Conference on Artificial Intelligence}}.
\newblock


\bibitem[Tian et~al\mbox{.}(2022)]%
        {tian2022gfsseg}
\bibfield{author}{\bibinfo{person}{Zhuotao Tian}, \bibinfo{person}{Xin Lai},
  \bibinfo{person}{Li Jiang}, \bibinfo{person}{Shu Liu},
  \bibinfo{person}{Michelle Shu}, \bibinfo{person}{Hengshuang Zhao}, {and}
  \bibinfo{person}{Jiaya Jia}.} \bibinfo{year}{2022}\natexlab{}.
\newblock \showarticletitle{Generalized Few-shot Semantic Segmentation}. In
  \bibinfo{booktitle}{\emph{CVPR}}.
\newblock


\bibitem[Tian et~al\mbox{.}(2019)]%
        {shapeawareembedding}
\bibfield{author}{\bibinfo{person}{Zhuotao Tian}, \bibinfo{person}{Michelle
  Shu}, \bibinfo{person}{Pengyuan Lyu}, \bibinfo{person}{Ruiyu Li},
  \bibinfo{person}{Chao Zhou}, \bibinfo{person}{Xiaoyong Shen}, {and}
  \bibinfo{person}{Jiaya Jia}.} \bibinfo{year}{2019}\natexlab{}.
\newblock \showarticletitle{Learning Shape-Aware Embedding for Scene Text
  Detection}. In \bibinfo{booktitle}{\emph{CVPR}}.
\newblock


\bibitem[Tian et~al\mbox{.}(2020)]%
        {pfenet}
\bibfield{author}{\bibinfo{person}{Zhuotao Tian}, \bibinfo{person}{Hengshuang
  Zhao}, \bibinfo{person}{Michelle Shu}, \bibinfo{person}{Zhicheng Yang},
  \bibinfo{person}{Ruiyu Li}, {and} \bibinfo{person}{Jiaya Jia}.}
  \bibinfo{year}{2020}\natexlab{}.
\newblock \showarticletitle{Prior Guided Feature Enrichment Network for
  Few-Shot Segmentation}.
\newblock \bibinfo{journal}{\emph{TPAMI}} (\bibinfo{year}{2020}).
\newblock


\bibitem[Vinyals et~al\mbox{.}(2016)]%
        {matchingnet}
\bibfield{author}{\bibinfo{person}{Oriol Vinyals}, \bibinfo{person}{Charles
  Blundell}, \bibinfo{person}{Tim Lillicrap}, \bibinfo{person}{Koray
  Kavukcuoglu}, {and} \bibinfo{person}{Daan Wierstra}.}
  \bibinfo{year}{2016}\natexlab{}.
\newblock \showarticletitle{Matching Networks for One Shot Learning}. In
  \bibinfo{booktitle}{\emph{NeurIPS}}.
\newblock


\bibitem[Wang et~al\mbox{.}(2019a)]%
        {wang2019panet}
\bibfield{author}{\bibinfo{person}{Kaixin Wang}, \bibinfo{person}{Jun~Hao
  Liew}, \bibinfo{person}{Yingtian Zou}, \bibinfo{person}{Daquan Zhou}, {and}
  \bibinfo{person}{Jiashi Feng}.} \bibinfo{year}{2019}\natexlab{a}.
\newblock \showarticletitle{Panet: Few-shot image semantic segmentation with
  prototype alignment}. In \bibinfo{booktitle}{\emph{ICCV}}.
\newblock


\bibitem[Wang et~al\mbox{.}(2018)]%
        {imaginary}
\bibfield{author}{\bibinfo{person}{Yu{-}Xiong Wang}, \bibinfo{person}{Ross~B.
  Girshick}, \bibinfo{person}{Martial Hebert}, {and} \bibinfo{person}{Bharath
  Hariharan}.} \bibinfo{year}{2018}\natexlab{}.
\newblock \showarticletitle{Low-Shot Learning From Imaginary Data}. In
  \bibinfo{booktitle}{\emph{CVPR}}.
\newblock


\bibitem[Wang et~al\mbox{.}(2019b)]%
        {wang2019dynamic}
\bibfield{author}{\bibinfo{person}{Yue Wang}, \bibinfo{person}{Yongbin Sun},
  \bibinfo{person}{Ziwei Liu}, \bibinfo{person}{Sanjay~E Sarma},
  \bibinfo{person}{Michael~M Bronstein}, {and} \bibinfo{person}{Justin~M
  Solomon}.} \bibinfo{year}{2019}\natexlab{b}.
\newblock \showarticletitle{Dynamic graph cnn for learning on point clouds}.
\newblock \bibinfo{journal}{\emph{TOG}} (\bibinfo{year}{2019}).
\newblock


\bibitem[Wu et~al\mbox{.}(2022)]%
        {wu2022point}
\bibfield{author}{\bibinfo{person}{Xiaoyang Wu}, \bibinfo{person}{Yixing Lao},
  \bibinfo{person}{Li Jiang}, \bibinfo{person}{Xihui Liu}, {and}
  \bibinfo{person}{Hengshuang Zhao}.} \bibinfo{year}{2022}\natexlab{}.
\newblock \showarticletitle{Point transformer V2: Grouped Vector Attention and
  Partition-based Pooling}. In \bibinfo{booktitle}{\emph{NeurIPS}}.
\newblock


\bibitem[Zermas et~al\mbox{.}(2017)]%
        {fastseg_autodrive}
\bibfield{author}{\bibinfo{person}{Dimitris Zermas}, \bibinfo{person}{Izzat
  Izzat}, {and} \bibinfo{person}{Nikolaos Papanikolopoulos}.}
  \bibinfo{year}{2017}\natexlab{}.
\newblock \showarticletitle{Fast segmentation of 3D point clouds: {A} paradigm
  on LiDAR data for autonomous vehicle applications}. In
  \bibinfo{booktitle}{\emph{ICRA}}.
\newblock


\bibitem[Zhang et~al\mbox{.}(2020)]%
        {deepemd}
\bibfield{author}{\bibinfo{person}{Chi Zhang}, \bibinfo{person}{Yujun Cai},
  \bibinfo{person}{Guosheng Lin}, {and} \bibinfo{person}{Chunhua Shen}.}
  \bibinfo{year}{2020}\natexlab{}.
\newblock \showarticletitle{DeepEMD: Few-Shot Image Classification With
  Differentiable Earth Mover's Distance and Structured Classifiers}. In
  \bibinfo{booktitle}{\emph{CVPR}}.
\newblock


\bibitem[Zhang et~al\mbox{.}(2019a)]%
        {canet}
\bibfield{author}{\bibinfo{person}{Chi Zhang}, \bibinfo{person}{Guosheng Lin},
  \bibinfo{person}{Fayao Liu}, \bibinfo{person}{Rui Yao}, {and}
  \bibinfo{person}{Chunhua Shen}.} \bibinfo{year}{2019}\natexlab{a}.
\newblock \showarticletitle{CANet: Class-Agnostic Segmentation Networks with
  Iterative Refinement and Attentive Few-Shot Learning}. In
  \bibinfo{booktitle}{\emph{CVPR}}.
\newblock


\bibitem[Zhang et~al\mbox{.}(2022)]%
        {mediseg}
\bibfield{author}{\bibinfo{person}{Dong Zhang}, \bibinfo{person}{Yi Lin},
  \bibinfo{person}{Hao Chen}, \bibinfo{person}{Zhuotao Tian},
  \bibinfo{person}{Xin Yang}, \bibinfo{person}{Jinhui Tang}, {and}
  \bibinfo{person}{Kwang{-}Ting Cheng}.} \bibinfo{year}{2022}\natexlab{}.
\newblock \showarticletitle{Deep Learning for Medical Image Segmentation:
  Tricks, Challenges and Future Directions}.
\newblock \bibinfo{journal}{\emph{CoRR}}  \bibinfo{volume}{abs/2209.10307}
  (\bibinfo{year}{2022}).
\newblock
\urldef\tempurl%
\url{https://doi.org/10.48550/arXiv.2209.10307}
\showDOI{\tempurl}
\showeprint[arXiv]{2209.10307}


\bibitem[Zhang et~al\mbox{.}(2019b)]%
        {hallucinate_saliency}
\bibfield{author}{\bibinfo{person}{Hongguang Zhang}, \bibinfo{person}{Jing
  Zhang}, {and} \bibinfo{person}{Piotr Koniusz}.}
  \bibinfo{year}{2019}\natexlab{b}.
\newblock \showarticletitle{Few-Shot Learning via Saliency-Guided Hallucination
  of Samples}. In \bibinfo{booktitle}{\emph{CVPR}}.
\newblock


\bibitem[Zhao et~al\mbox{.}(2019)]%
        {zhao2019pointweb}
\bibfield{author}{\bibinfo{person}{Hengshuang Zhao}, \bibinfo{person}{Li
  Jiang}, \bibinfo{person}{Chi-Wing Fu}, {and} \bibinfo{person}{Jiaya Jia}.}
  \bibinfo{year}{2019}\natexlab{}.
\newblock \showarticletitle{{PointWeb}: Enhancing Local Neighborhood Features
  for Point Cloud Processing}. In \bibinfo{booktitle}{\emph{CVPR}}.
\newblock


\bibitem[Zhao et~al\mbox{.}(2021b)]%
        {zhao2021point}
\bibfield{author}{\bibinfo{person}{Hengshuang Zhao}, \bibinfo{person}{Li
  Jiang}, \bibinfo{person}{Jiaya Jia}, \bibinfo{person}{Philip~HS Torr}, {and}
  \bibinfo{person}{Vladlen Koltun}.} \bibinfo{year}{2021}\natexlab{b}.
\newblock \showarticletitle{Point transformer}. In
  \bibinfo{booktitle}{\emph{CVPR}}.
\newblock


\bibitem[Zhao et~al\mbox{.}(2021a)]%
        {zhao2021few}
\bibfield{author}{\bibinfo{person}{Na Zhao}, \bibinfo{person}{Tat-Seng Chua},
  {and} \bibinfo{person}{Gim~Hee Lee}.} \bibinfo{year}{2021}\natexlab{a}.
\newblock \showarticletitle{Few-shot 3d point cloud semantic segmentation}. In
  \bibinfo{booktitle}{\emph{CVPR}}.
\newblock


\bibitem[Zhao et~al\mbox{.}(2023)]%
        {zhao2023towards}
\bibfield{author}{\bibinfo{person}{Zicheng Zhao}, \bibinfo{person}{Linhao Luo},
  \bibinfo{person}{Shirui Pan}, \bibinfo{person}{Quoc Viet~Hung Nguyen}, {and}
  \bibinfo{person}{Chen Gong}.} \bibinfo{year}{2023}\natexlab{}.
\newblock \showarticletitle{Towards Few-shot Inductive Link Prediction on
  Knowledge Graphs: A Relational Anonymous Walk-guided Neural Process
  Approach}.
\newblock \bibinfo{journal}{\emph{ECML-PKDD}} (\bibinfo{year}{2023}).
\newblock


\end{thebibliography}

% If your work has an appendix, this is the place to put it.
\clearpage
\appendix
\begin{table}[!t]
\begin{tabular}{c|c|c}
\toprule
& split=0  & split=1  \\
\hline
\textbf{S3DIS} &
  \begin{tabular}[c]{@{}c@{}}beam, board, bookcase,\\ ceiling, chair, column\end{tabular} &
  \begin{tabular}[c]{@{}c@{}}door, floor, sofa,\\ table, wall, window\end{tabular} \\ \hline
\textbf{ScanNet} &
  \begin{tabular}[c]{@{}c@{}}otherfurniture, picture,\\ refrigerator, show curtain,\\ sink, sofa, table,\\ toilet, wall, window\end{tabular} &
  \begin{tabular}[c]{@{}c@{}}bathtub, bed, bookshelf,\\ cabinet, chair, counter, \\ curtain, desk, door, floor\end{tabular}
   \\
\bottomrule
\vspace{-0.1cm}
\end{tabular}

\caption{Test class names for each split of S3DIS and ScanNet.}
\label{tab:split_details}
% \vspace{-0.4cm}
\end{table}

\section{Dataset Split}
\label{sec:dataset_split}

\mypara{Details of Splits.} The details of test class names in each split of the S3DIS and ScanNet are shown in Table~\ref{tab:split_details}.

\section{Additional Ablation Study}
\label{sec:additional_ablation_study}

\mypara{Designs for correlation module in BPA. } 
In Section~\ref{sec:bpa}, we introduced the use of module $\mathcal{C}$ to extract $\mathcal{r}$ for adapting the support background prototype, as defined by $\mathcal{r} = \mathcal{C}(\vec{p}_s^{bg}, \vec{f}_q, \vec{p}_s^{fg})$. Here, we conducted experiments to compare the performance of using cross-attention for $\mathcal{C}$ against two alternatives: applying $\mathcal{C}$ as cross-attention to $\vec{p}_s^{bg}$ and $\vec{f}_q$, without considering $\vec{p}_s^{fg}$ (experiment (a)), and simply adopting an MLP as $\mathcal{C}$ (experiment (b)). As shown in Table \ref{tab:different_prototype_denoising_design}, both alternatives resulted in significant performance degradation compared to our proposed design (experiment (c)) mentioned in Section~\ref{sec:bpa}.

\begin{table}[!t]
\tabcolsep=0.75cm
\begin{tabular}{lcc}
\toprule
\multicolumn{2}{l}{Designs for correlation module}                 & \multicolumn{1}{c}{mIoU(\%)}                \\
\midrule
\multicolumn{2}{l}{(a)  $\mathcal{C}(\vec{p}_s^{bg}, \vec{f}_q) =$ Cross-Attention}      & 58.70                   \\
\multicolumn{2}{l}{(b)  $\mathcal{C}(\vec{p}_s^{bg}, \vec{f}_q, \vec{p}_s^{fg}) = $ MLP}  & 47.46                   \\
\multicolumn{2}{l}{(c)  Our design}                                                      & \textbf{71.36}                  \\
\bottomrule
\end{tabular}
\caption{The effects of different designs for BPA.}
\label{tab:different_prototype_denoising_design}
\end{table}

\mypara{Alternative designs for support foreground information filtering in BPA. } 
In Table~\ref{tab:strategies_for_fg_filter}, we further evaluate several strategies for filtering foreground (FG) information, specifically: (a) direct subtraction of $\mathcal{r}_s$ from $\mathcal{r}_q$, i.e., $\mathcal{r} = \mathcal{r}_q - \mathcal{r}_s$; (b) employing a learnable projector to automatically learn this strategy, i.e., $\mathcal{r} = \mathcal{P}(\mathcal{r}_q, \mathcal{r}_s)$; and (c) our proposed design, which leverages a mask generated by an MLP layer followed by a $sigmoid$ function to suppress FG information in $\mathcal{r}_q$, as outlined in Section 3.2 Eq.~\eqref{eq:FG_information_filter}: $\mathcal{r} = \mathcal{G}( \mathcal{r}_q,  \mathcal{r}_s) = \mathcal{r}_q \cdot ( 1 - \sigma(\mathcal{r}_s))$. 

Our experimental results in Table~\ref{tab:strategies_for_fg_filter} demonstrate that our proposed method achieves the most favorable performance by effectively extracting $\mathcal{r}$ to adapt support background prototypes to the query background without compromising the distinction between foreground and background, as compared to (a) and (b).

\begin{table}[t]
\tabcolsep=0.7cm
\begin{tabular}{lcc}
\toprule
\multicolumn{2}{l}{Different strategies for FG filtering}                 & \multicolumn{1}{c}{mIoU(\%)}  \\
\midrule
\multicolumn{2}{l}{(a) $\mathcal{r} = \mathcal{r}_q - \mathcal{r}_s$}                       & 63.72           \\
\multicolumn{2}{l}{(b) $\mathcal{r} = \mathcal{P}(\mathcal{r}_q, \mathcal{r}_s)$}           & 48.12           \\
\multicolumn{2}{l}{(c) $\mathcal{r} = \mathcal{r}_q \cdot ( 1 - \sigma(\mathcal{r}_s))$}    & \textbf{71.36}           \\
\bottomrule
\end{tabular}
\caption{The effects of different strategies for FG information filtering in BPA.}
\label{tab:strategies_for_fg_filter}
\vspace{-0.3cm}
\end{table}

\mypara{Temperature values for scaling cosine similarity. } As mentioned in the main paper, in order to make the cosine outputs of ProtoNet can be decently optimized by the cross-entropy loss, a temperature factor $\vec{t}_{CE}$ is used to scale the value range of cosine $\theta$ from $[-1,1]$ to $[-\vec{t}_{CE}, \vec{t}_{CE}]$. To investigate the impacts brought by different values of $\vec{t}_{CE}$, we conduct experiments on S3DIS under the 1-way 1-shot $S^0$ setting, and the results are shown in Table \ref{tab:t_of_cosine_pred}. We observed that setting $\vec{t}_{CE}$ to 15 can bring a decent performance.

\begin{table}[h]
\begin{tabular}{c|cccc}
\toprule
$\vec{t}_{CE}$            & 1        & 5          & 15                  & 30     \\
\midrule
mIoU                 & 62.56    & 64.87      & \textbf{66.18}      & 64.91  \\
\bottomrule
\end{tabular}
\caption{The results of different values of $\vec{t}_{CE}$.}
\label{tab:t_of_cosine_pred}
\vspace{-0.3cm}
\end{table}

\mypara{Temperature values for distillation. } The temperature factor $\vec{t}_{KL}$ in the KL divergence of Holistic Rectification (HR) can be utilized to control the degree of smoothness of knowledge during distillation. We conducted experiments on S3DIS under the 1-way 1-shot $S^0$ setting to investigate this, and the results are presented in Table~\ref{tab:t_of_KL}. We observed that setting $\vec{t}_{KL}$ to 1 can lead to better performance with HR.

\begin{table}[h]
\begin{tabular}{c|ccccc}
\toprule
$\vec{t}_{KL}$            & 0.1      & 0.5      & 1                & 5         \\
\midrule
mIoU                      & 67.19    & 69.51    & \textbf{69.54}   & 65.17     \\
\bottomrule
\end{tabular}
\caption{The results of different values of $\vec{t}_{KL}$.}
\label{tab:t_of_KL}
% \vspace{-0.4cm}
\end{table}

\end{document}

\title{Boosting Few-shot 3D Point Cloud Segmentation via Query-Guided Enhancement \\ Supplementary Material}

\maketitle

\vspace{0.5cm}
\section*{Overview}
\vspace{0.3cm}
This is the supplementary file for our submission titled \textit{Boosting Few-shot 3D Point Cloud Segmentation via Query-Guided Enhancement}.
This material supplements the main paper with the following content:
\vspace{0.3cm}
\begin{itemize}
	\item (\ref{sec:more_framework_details}) \textbf{MORE FRAMEWORK DETAILS}.
            \vspace{0.2cm}
            \iitem a. Framework of BPA.
            \vspace{0.2cm}
            \iitem b. Extend our method to n-way k-shot multi-prototype.
            \vspace{0.2cm}
            \iitem c. Label propagation.
        \vspace{0.3cm}
	\item (\ref{sec:dataset_split}) \textbf{DATASET SPLIT}.
        \vspace{0.3cm}
	\item (\ref{sec:additional_ablation_study}) \textbf{ADDITIONAL ABLATION STUDY}.
            \vspace{0.2cm}
            \iitem a. Temperature values for scaling cosine similarity.
            \vspace{0.2cm}
            \iitem b. Temperature values for distillation.
        \vspace{0.3cm}
	\item (\ref{sec:more_qualitative_results}) \textbf{MORE QUALITATIVE RESULTS}.
            \vspace{0.2cm}
            \iitem a. More t-SNE results.
            \vspace{0.2cm}
            \iitem b. More prediction visualizations.
\end{itemize}

\newpage

\section{More Framework Details}
\label{sec:more_framework_details}

% If your work needs an appendix, add it before the
% ``\verb|\end{document}|'' command at the conclusion of your source
% document.

% Start the appendix with the ``\verb|appendix|'' command:
% \begin{verbatim}
%   \appendix
% \end{verbatim}
% and note that in the appendix, sections are lettered, not
% numbered. This document has two appendices, demonstrating the section
% and subsection identification method.

\begin{figure}[h]
  \centering
  \includegraphics[width=\linewidth]{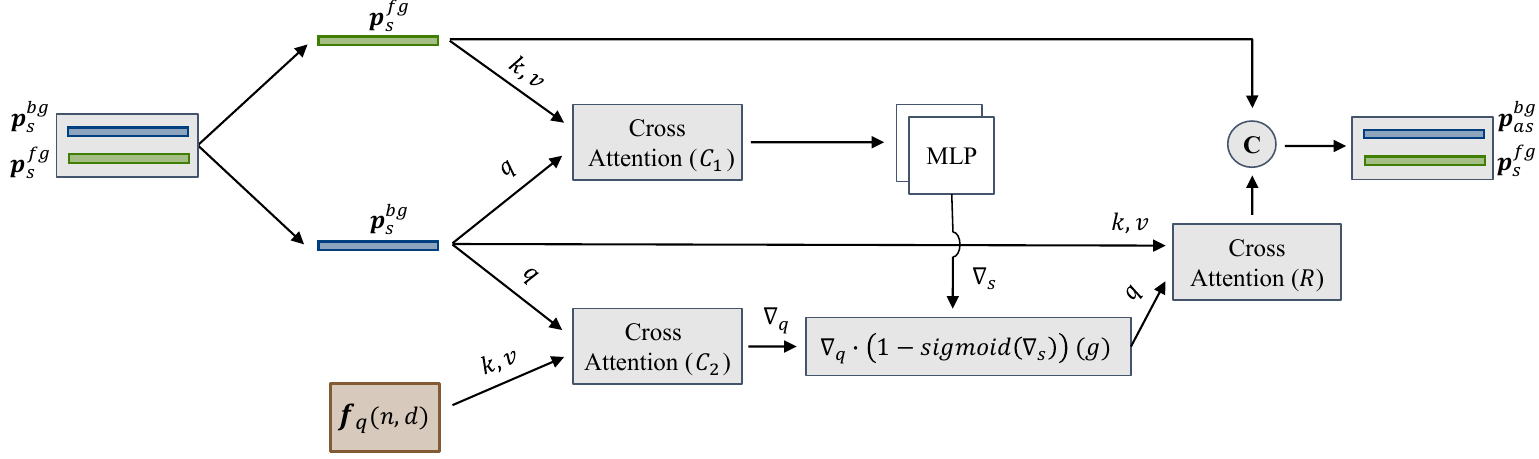}
  \caption{The framework details of Background Prototype Adaptation, where $\vec{C}$ represents $concat$ and all the Cross-Attention modules are shared weights.}
  \Description{}
  \label{fig:BPA}
  % \vspace{0.2cm}
\end{figure}

% \mypara{Framework of BPA.} Figure \ref{fig:BPA} illustrated the framework details of Background Prototype Adaptation. BPA contains three Cross-Attention modules and a 2-layer MLP. Notably, in order to guarantee that $\vec{f}_{q}$ and $\vec{p}_{s}$ interact in the same feature space, all the three Cross-Attention modules are shared weights.

\mypara{Framework of BPA.} Figure 1 depicts the framework details of Background Prototype Adaptation (BPA), which consists of three Cross-Attention modules and a two-layer MLP. It is worth noting that to ensure the interaction between $\vec{f}_{q}$ and $\vec{p}_{s}$ in the same feature space, all three Cross-Attention modules share the same weights.

\mypara{Extend our method to n-way k-shot multi-prototype.} In the main paper, we illustrate our idea with the single-prototype example for facilitating understanding. In the following, we detail the way for adapting our method to the n-way k-shot multi-prototype pipeline such as AttMPTI. 

Firstly, the Farthest Point Search (FPS) algorithm is applied to query features $\left\{\vec{f}_{q}^{i}\right\}_{i=1}^{c}$ to get seeds $\left\{\vec{s}^{i}\right\}_{i=1}^{c}$, where $c$ represents the total number of categories, including both n-way categories and the background category. Notably, the processes for different categories are independent. In the k-shot setting, for each category, we first merge k support point clouds into one point cloud to obtain multi-prototypes for the subsequent process. Secondly, the distances between $\vec{s}^{i}$ and $\vec{f}_{q}^{i}$ are calculated by L2 distance. Then, each query feature of $\vec{f}_{q}^{i}$ is assigned to the closest seed. Finally, the average of the features assigned to each seed is used as its prototype. 

Take 2-way 5-shot setting as an example, our method can be easily extended to the n-way k-shot multi-prototype setting given support multi-prototype $\vec{p}_s$. Firstly, we obtain the query feature $\vec{f}_{q} \in \Omega^{[2 \times n \times d]}$ from the feature extractor and support multi-prototype $\vec{p}_{s} \in \Omega^{[(2+1) \times m \times d]}$ from multi-prototype generation, where $n$, $m$ and $d$ denote the number of points in a point cloud, the number of multi-prototype for each class, and the feature dimension channels. Secondly, we reshape $\vec{f}_{q} \in \Omega^{[2 \times n \times d]}$ and $\vec{p}_{s} \in \Omega^{[(2+1) \times m \times d]}$ to $\vec{f}_{q} \in \Omega^{[2n \times d]}$ and $\vec{p}_{s} \in \Omega^{[(2+1)m \times d]}$. Then, we can treat this scenario as a 1-way 1-shot situation in our method, and subsequent processing is performed as described in the main paper.

% Formally, the multi-prototype $\vec{p}^{k}$ of class k are given by:
% \begin{equation}
% \begin{split}
%     &\vec{p}^{k} = \left\{ \vec{p}_{1}^{k} \dots \vec{p}_{u}^{k}|\vec{p}_{i}^{k} = \frac{1}{|\mathcal{I}^{k\ast}|} \sum_{\vec{f}_{j}^{k} \in \mathcal{I}^{k\ast}}^{} \vec{f}_{j}^{k} \right\} \\
%     &s.t. \quad \arg\min_{\mathcal{I}^{k}} \sum_{i=1}^{u} \sum_{\vec{f}_{j}^{k} \in \mathcal{I}_{i}^{k}}^{}\left \|\vec{f}_{j}^{k} - \vec{s}_{i}^{k} \right \|_{2}
% \end{split}
% \end{equation}
% where $\left\{ \vec{f}_{j}^{k} \right\}_{j=1}^{v}$ are partitioned into $u$ sets $\mathcal{I}^{k\ast} = \{\mathcal{I}_{1}^{k\ast} \dots \mathcal{I}_{u}^{k\ast}\}$, such that $\vec{f}_{j}^{k} \in \mathcal{I}_{i}^{k\ast}$ is assigned to $\vec{s}_{i}^{k}$.

% \zttian{The detailed generation of multiple prototypes can be simplified, such as this equation, because details can be found in other papers. What I want to add is how to adapt our method to the multi-prototype pipeline, instead of the introduction of multi-prototype generation.}

\mypara{Label propagation.} The prediction generator in AttMPTI is label propagation, we will give it a brief introduction here, and please check~\cite{zhao2021few} for more details. 

Once multiple prototypes are obtained, we utilize them and query features $\vec{f}_q$ as nodes to construct a $k$-NN graph $\vec{S}$. Furthermore, a label matrix $\vec{Y}$ is also defined, where c represents the number of categories. The rows in this matrix correspond to multi-prototype are one-hot ground truth labels and the rests are zero. It is important to note that during testing, only the labels of the support set are available. Thus, only the multi-prototypes produced by the support features $\vec{f}_s$ have labels in the label matrix, while the remaining entries are set to zero. Given $\vec{S}$ and $\vec{Y}$, we can iteratively propagate labels from support to query through the entire graph according to:
\begin{equation}
    \vec{Z_{t+1}} = \alpha\vec{S}\vec{Z}_t + (1-\alpha)\vec{Y}.
\end{equation}
And \cite{NIPS2003_87682805} demonstrates that the above equation has a closed-form solution:
\begin{equation}
    \vec{Z}^{\ast} = (\vec{I} - \alpha\vec{S})^{-1}\vec{Y}.
\end{equation}
The closed-form solution's $\vec{Z}^{\ast}$ is utilized as the prediction results in AttMPTI.

\section{Dataset Split}
\label{sec:dataset_split}

\begin{table}[t]
\tabcolsep=0.75cm
\begin{tabular}{c|c|c}
\toprule
& split=0  & split=1  \\
\hline
\textbf{S3DIS} &
  \begin{tabular}[c]{@{}c@{}}beam, board, bookcase,\\ ceiling, chair, column\end{tabular} &
  \begin{tabular}[c]{@{}c@{}}door, floor, sofa,\\ table, wall, window\end{tabular} \\ \hline
\textbf{ScanNet} &
  \begin{tabular}[c]{@{}c@{}}otherfurniture, picture,\\ refrigerator, show curtain,\\ sink, sofa, table,\\ toilet, wall, window\end{tabular} &
  \begin{tabular}[c]{@{}c@{}}bathtub, bed, bookshelf,\\ cabinet, chair, counter, \\ curtain, desk, door, floor\end{tabular}
   \\
\bottomrule
\end{tabular}
\caption{Test class names for each split of S3DIS and ScanNet.}
\label{tab:split_details}
% \vspace{-0.4cm}
\end{table}

\mypara{Details of Splits.} The details of test class names in each split of the S3DIS and ScanNet are shown in Table~\ref{tab:split_details}.

\section{Additional Ablation Study}
\label{sec:additional_ablation_study}

\begin{table}[h]
\tabcolsep=1cm
\begin{tabular}{c|cccc}
\toprule
$\vec{t}_{CE}$            & 1        & 5          & 15                  & 30     \\
\midrule
mIoU                 & 62.56    & 64.87      & \textbf{66.18}      & 64.91  \\
\bottomrule
\end{tabular}
\caption{The results of different values of $\vec{t}_{CE}$.}
\label{tab:t_of_cosine_pred}
% \vspace{-0.4cm}
\end{table}

\mypara{Temperature values for scaling cosine similarity. } As mentioned in the main paper, in order to make the cosine outputs of ProtoNet can be decently optimized by the cross-entropy loss, a temperature factor $\vec{t}_{CE}$ is used to scale the value range of cosine $\theta$ from $[-1,1]$ to $[-\vec{t}_{CE}, \vec{t}_{CE}]$. To investigate the impacts brought by different values of $\vec{t}_{CE}$, we conduct experiments on S3DIS under the 1-way 1-shot $S^0$ setting, and the results are shown in Table \ref{tab:t_of_cosine_pred}. We observed that setting $\vec{t}_{CE}$ to 15 can bring a decent performance.

\begin{table}[h]
\tabcolsep=1cm
\begin{tabular}{c|ccccc}
\toprule
$\vec{t}_{KL}$            & 0.1      & 0.5      & 1                & 5         \\
\midrule
mIoU                      & 67.19    & 69.51    & \textbf{69.54}   & 65.17     \\
\bottomrule
\end{tabular}
\caption{The results of different values of $\vec{t}_{KL}$.}
\label{tab:t_of_KL}
% \vspace{-0.4cm}
\end{table}

\mypara{Temperature values for distillation. } The temperature factor $\vec{t}_{KL}$ in the KL divergence of Holistic Rectification (HR) can be utilized to control the degree of smoothness of knowledge during distillation. We conducted experiments on S3DIS under the 1-way 1-shot $S^0$ setting to investigate this, and the results are presented in Table~\ref{tab:t_of_KL}. We observed that setting $\vec{t}_{KL}$ to 1 can lead to better performance with HR.

\section{More Qualitative Results}
\label{sec:more_qualitative_results}

\begin{figure*}[t]
    \centering
    \includegraphics[width=\textwidth]{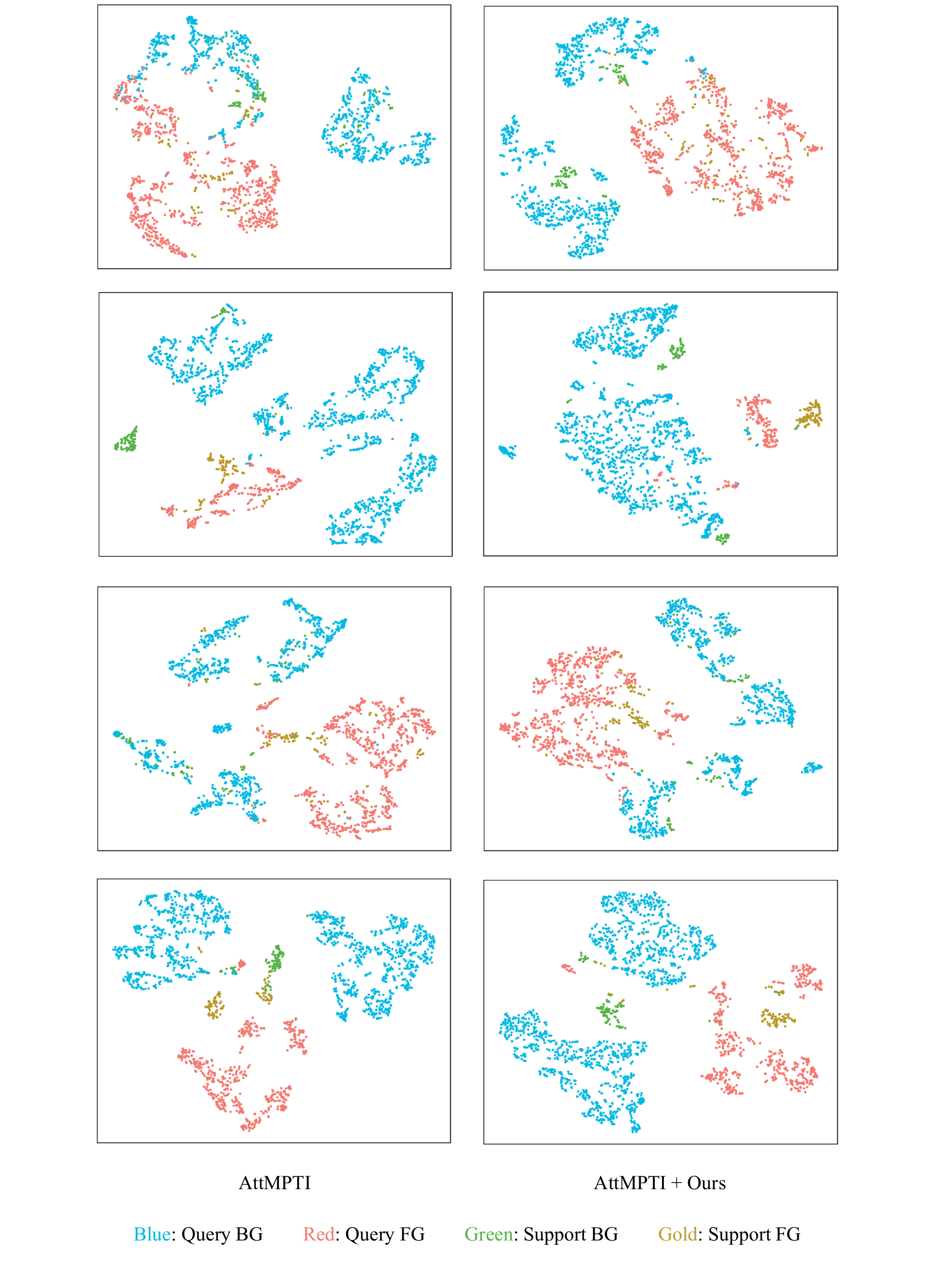}
    \caption{More t-SNE results of query features and support prototypes in AttMPTI and our method. The query background and foreground features are denoted by blue and red, respectively, while the support background and foreground prototypes are depicted using green and gold.}
    \label{fig:tsne2}
\end{figure*}

\mypara{More t-SNE results.} Figure~\ref{fig:tsne2} displays additional t-SNE results of the query features and support prototypes in both AttMPTI and our proposed method. The various colors in the results signify different elements, namely the query background features, query foreground features, support background prototypes, and support foreground prototypes. The results show that the support prototypes in our method are more discriminative.

\begin{figure*}[t]
    \centering
    \includegraphics[width=\textwidth]{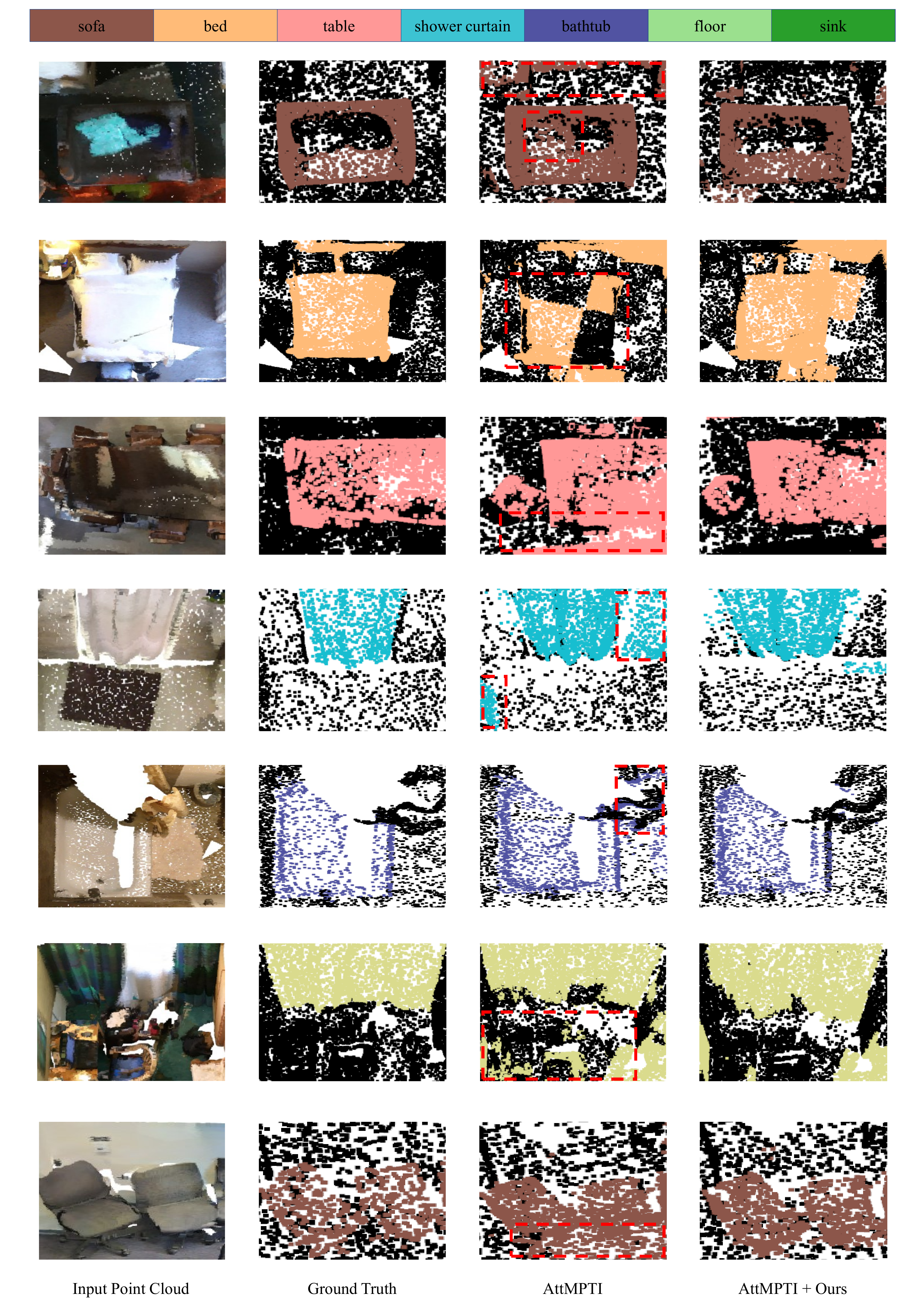}
    \caption{More visualizations of the prediction results.}
    \label{fig:pred_visualization2}
\end{figure*}

\begin{figure*}[t]
    \centering
    \includegraphics[width=\textwidth]{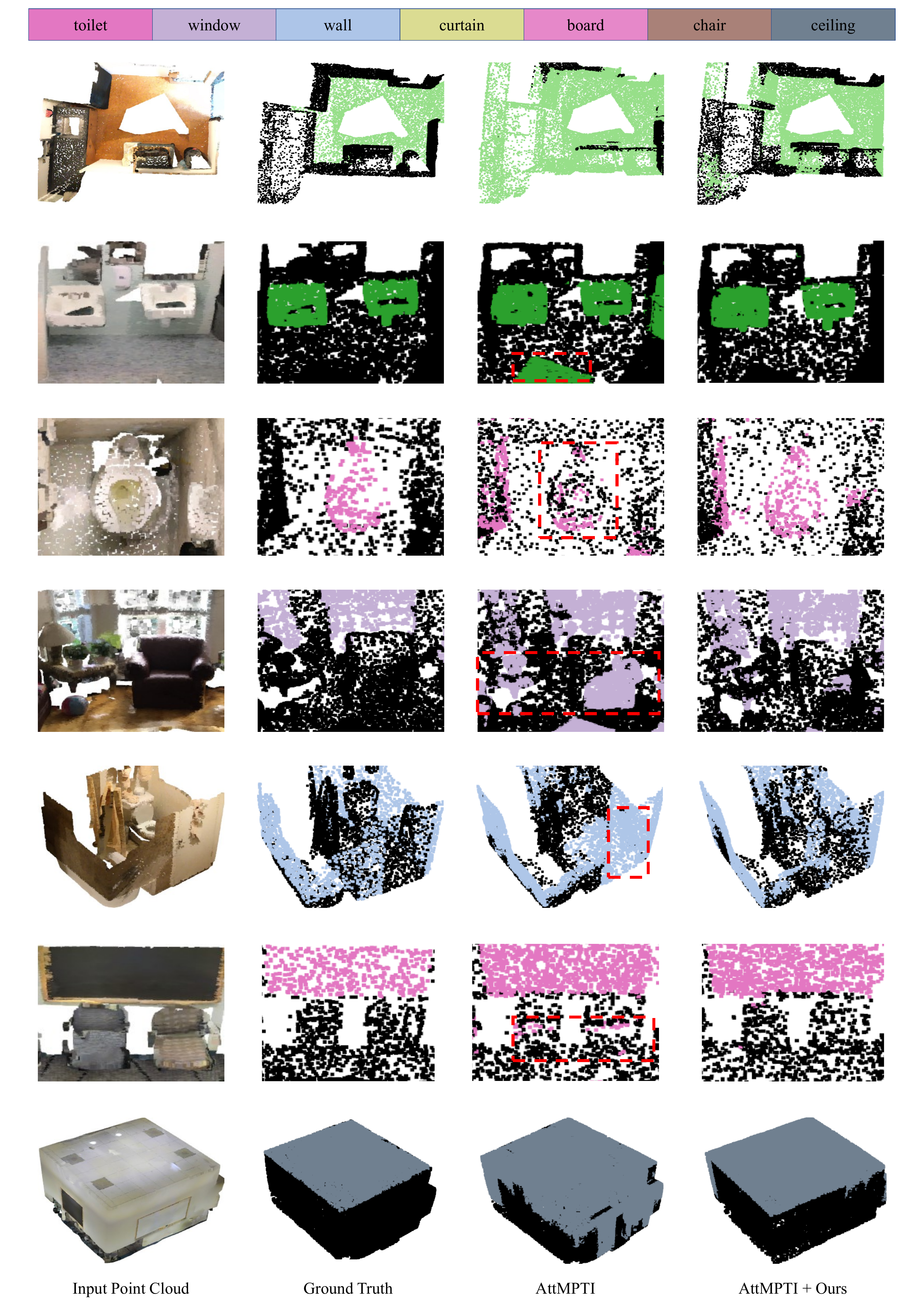}
    \caption{More visualizations of the prediction results.}
    \label{fig:pred_visualization3}
\end{figure*}

\mypara{More prediction visualization.} To further demonstrate the effectiveness of our method for various categories and scenes, we provide additional prediction visualizations in Figure~\ref{fig:pred_visualization2} and Figure~\ref{fig:pred_visualization3}. In these visualizations, we compare the input point cloud, ground truth, AttMPTI, and our method based on AttMPTI.

\bibliographystyle{ACM-Reference-Format}
\bibliography{base}